\documentclass[11pt]{article}
\usepackage{acl}
\usepackage{orcidlink}

\usepackage{times}
\usepackage{latexsym}

\usepackage{xcolor}
\usepackage{colortbl}
\usepackage[T1]{fontenc}
\usepackage{inconsolata}
\usepackage{fontawesome5}

\usepackage[utf8]{inputenc}

\usepackage{microtype}

\usepackage{inconsolata}
\usepackage{hyphenat}

\usepackage{graphicx}

\usepackage[T1]{fontenc}
\usepackage[utf8]{inputenc}
\usepackage{microtype}
\usepackage{inconsolata}
\usepackage{todonotes}
\usepackage{times}
\usepackage{graphicx}
\usepackage{booktabs}
\usepackage{amsmath}
\usepackage{amssymb}
\usepackage{amsfonts}
\usepackage{multirow}
\usepackage{array}
\usepackage{enumitem}
\usepackage{algorithm}
\usepackage[most]{tcolorbox}
\usepackage{algpseudocode}

\hypersetup{
  colorlinks=true,
  linkcolor=black!50!black,
  citecolor=black!50!black,
  urlcolor=black!70!black
}

\newcommand{\method}{\texttt{Tree-of-Concerns}}
\newcommand{\methodshort}{\texttt{ToC}}
\newcommand{\bench}{\texttt{ToC-Bench}}
\newcommand{\panel}{\texttt{Panel Review}}
\newcommand{\todp}{\texttt{Tree-of-Debate}}

\title{\method{}: Hierarchical Multi-Agent Debate for Unstated-Limitation Extraction in Scientific Critique}

\author{
  Sahil Mishra \orcidlink{0000-0001-5477-9003}\hspace{2.5em}
  Niranjan Rajeev \orcidlink{0009-0004-4309-5537}\hspace{2.5em} 
  Tanmoy Chakraborty \orcidlink{0000-0002-0210-0369} \\
   Indian Institute of Technology Delhi, India \\
  \texttt{\{sahil.mishra, niranjan.rajeev, tanchak\}@ee.iitd.ac.in}
}

\date{}

\begin{document}

\maketitle

\begin{abstract}
As scientific literature grows and papers increasingly under-report limitations, multi-agent LLMs offer a promising approach to systematically uncover these hidden failure modes. Here, we introduce \method{}, a multi-agent framework that deploys specialized skeptic personas, each operating through a category-specific analytical lens, as parallel debate trees to extract unstated limitations from scientific papers. Each persona conducts structured, evidence-grounded argumentation, while a \panel{} mechanism re-evaluates each surviving claim from all five perspectives to correct category drift and severity miscalibration. Through retrieval-free, single-paper experiments on \bench{}, our benchmark of 414 research papers with 1{,}905 unstated limitations, sourced from reviewer-reported weaknesses and follow-up citation critiques, we demonstrate that \methodshort{} improves precision by 79\% and coverage by 11\% relative to the strongest baseline, surfacing specific, evidence-grounded concerns that support reviewers in systematic evaluation.
\end{abstract}

\section{Introduction}
\label{sec:intro}
Scientific papers systematically under-report their limitations due to incentive misalignment, page constraints, and authorial blind spots \citep{Ioannidis2007, Vazire2017}, with a recent analysis finding 73\% of reviewer-identified weaknesses absent from NeurIPS 2023 limitation sections \citep{Liangetal2024}. This gap places an unsustainable burden on peer reviewers who must independently identify these unacknowledged issues under deadlines. However, the consequences extend beyond the review process. When limitations remain unstated, practitioners risk deploying methods outside their valid operating range, leading to silent production failures and contributing to a broader reproducibility crisis within machine learning \citep{Kapoor2024, semmelrock2025reproducibility}. Consequently, a systematic method for surfacing these hidden limitations is critical to support both the peer review process and the broader downstream research and deployment pipeline.

\begin{figure}[!t]
 \centering
 \includegraphics[width=0.99\linewidth]{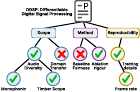}
 \caption{A taxonomy of limitations made for Paper $P$, specific to \textit{Scope}, \textit{Method} and \textit{Reproducibility}. Green check mark: unstated limitation; red cross: limitation stated by user.}
 \label{fig:intro}
\end{figure}

To address this bottleneck, automated scientific critique has emerged as a promising direction. Systems such as DIAGPaper \citep{diagpaper2024}, MARG \citep{damico2024marg}, and AgentReview \citep{jin2024agentreview} deploy LLM-based agents to generate review-like feedback. However, these systems are designed to \emph{replicate} the style and coverage of human reviews rather than to systematically extract limitations that authors have failed to state (cf. Fig. \ref{fig:intro}). This distinction is critical because a system optimized to match human review distributions will naturally gravitate toward the same salient issues that reviewers mention. Furthermore, recent studies reveal that standard multi-agent consensus mechanisms often suffer from inherent conformity, where agents are overly influenced by the majority and propagate errors rather than engaging in rigorous, independent verification \cite{Cuietal2025, Wengetal2025, Choietal2025}. Without specialized analytical lenses and mechanisms to preserve viewpoint diversity, naive agent swarms are prone to premature homogenization and groupthink \cite{WuIto2025}, failing to surface deeper, more specific concerns. We focus on retrieval-free, single-paper identification of internal scientific limitations, rather than general-purpose peer review. Writing assessment, novelty evaluation, and missing-citation detection are outside our scope because they require different evidence or external retrieval.

Two fundamental gaps currently hinder the shift from review replication to genuine limitation discovery. First, although recent work studies limitation identification and generation \citep{bagels2025,limagents2026,limitgen2025}, existing resources differ in whether they include author-acknowledged limitations, synthetic perturbations, or external retrieval. General review datasets such as PeerRead \citep{kang2018dataset}, NLPeer \citep{dycke2023nlpeer}, and AAAR-1.0 \citep{aaar2024} also do not isolate externally grounded, author-unacknowledged limitations. Second, existing multi-agent frameworks do not combine retrieval-free single-paper critique, independent category-specialized branches, and post-hoc cross-agent reconciliation. While \todp{} \citep{kargupta2025tod} demonstrates the value of debate, its homogeneous agents operate on pairwise comparisons and can converge on obvious flaws rather than exploring diverse, complex failure modes. To bridge these gaps, we propose three design principles:

\textbf{Specialized personas prevent critique groupthink.} When prompted to identify limitations without constraints, generalist LLMs gravitate toward the same salient methodological issues that authors naturally self-report. We explore the use of specialized skeptic personas, each constrained to a distinct analytical lens (scope, methodology, theoretical, reproducibility, fairness). By forcing agents to evaluate a paper through a narrow failure-mode constraint, we break the gravitational pull of stated limitations and foster the discovery of genuinely novel, unstated concerns.

\textbf{Tree-structured debate stress-tests and deepens claims.} A single pass of generation is insufficient for rigorous critique, as LLMs frequently hallucinate missing baselines or raise easily refutable concerns \citep{huang2025survey}. We propose converting limitation discovery into a tree-structured adversarial debate. Within each node, a skeptic's claim is attacked by a ``Paper Advocate'' agent, forcing the skeptic to withdraw invalid claims or revise them with stronger evidence. If a claim survives and suggests a deeper structural issue, a moderator expands it into a child node, allowing for depth-first exploration of complex, multi-layered flaws.

\textbf{Cross-branch reconciliation resolves category drift.} While isolating specialized branches prevents premature consensus, it introduces new failure modes: cross-branch redundancy (e.g., the Scope and Methodology skeptics finding the same issue from different angles) and category drift (e.g., a theoretical skeptic mislabeling a scope issue as a proof gap). Because isolated trees lack global context, we propose a post-hoc panel review mechanism. This panel evaluates the aggregate critique holistically, labeling cross-category overlap, recalibrating inflated severities, and reclassifying drifted categories to produce a unified, coherent output.

We integrate these principles into \method{}, a novel framework that dynamically deploys five specialized parallel debate trees. Each persona conducts a structured four-stage argumentation loop (claim, advocate response, revision, moderation) grounded by direct paper evidence. To evaluate retrieval-free, single-paper unstated-limitation extraction, we pair our framework with \bench{}.

Our contributions are summarized as follows:\footnote{\href{https://github.com/sahilmishra0012/Tree-of-Concerns} {Code: \faGithub\ sahilmishra0012/Tree-of-Concerns}}
\begin{itemize}[leftmargin=*, noitemsep, topsep=0pt]
\item We introduce \bench{}, a benchmark specifically curated for retrieval-free evaluation of author-unacknowledged limitations. It contains 414 papers with 1{,}905 externally sourced reference limitations from peer-review weaknesses and follow-up citation critiques.
\item We propose \method{}, a hierarchical multi-agent framework that pairs specialized, parallel debate trees with an adversarial per-node filtering mechanism to extract deep, evidence-grounded critiques.
\item We introduce \panel{}, a cross-branch reconciliation mechanism that holistically reviews parallel outputs to record cross-category applicability, correct category drift, and calibrate severity.
\item Through rigorous evaluation, we demonstrate that \method{} achieves 36.1\% Coverage@10 and 40.3\% Precision on \bench{}, showing improvements of 79\% in precision and 11\% in coverage over the strongest baseline.
\end{itemize}

\section{Related Work}
\label{sec:relwork}

\subsection{Multi-Persona LLM Debate}
Multi-agent debate frameworks use structured disagreement to enhance reasoning and reduce hallucinations \citep{du2023improving, liang2023encouraging, wang2024unleashing}. However, open multi-agent debate can also induce conformity and degrade performance \citep{wynn2025}. Building on the ``Society of Mind'' tradition \citep{zhuge2024language}, existing systems typically assign stance-based roles to adjudicate claims \citep{irving2018ai, chan2023chateval} or compare papers pairwise, as in \todp{} \citep{kargupta2025tod}. In contrast, \method{} isolates category-specialized branches during generation and uses the Advocate as an adversarial filter rather than as a consensus-seeking participant. Other works use LLMs for taxonomy expansion \citep{mishra2024flame,mishra2025lorex} or develop representations for hierarchical relations \citep{mishra2026quantaxo,mishra2026taxobell,mishra2026polaris}; \method{} instead uses a hierarchy to organize and refine concerns about one paper.

\begin{figure*}
 \centering
 \includegraphics[width=0.9\linewidth]{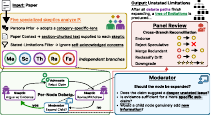}
\caption{\textbf{The \method{} framework.} Five parallel skeptic branches stress-test candidate limitations via adversarial debate; the \panel{} then resolves cross-branch redundancy, category drift, and severity miscalibration.}
\label{fig:main_architecture}
\end{figure*}

\subsection{Automated Paper Critique}
Recent LLM frameworks automate scientific critique by simulating peer-review dynamics through reviewer--author dialogues, meta-reviews, and structured prompting \citep{yuan2022can, tyser2024reviewergpt, damico2024marg, jin2024agentreview, diagpaper2024}. Because these systems generally target broad review generation, their outputs may emphasize salient concerns already acknowledged by authors \citep{wang2023large, Liangetal2024}. \method{} instead targets retrieval-free identification of author-unacknowledged limitations using isolated, specialized adversarial debate trees.

\subsection{Benchmarks for Scientific Text}
Existing benchmarks for review generation \citep{kang2018dataset, dycke2023nlpeer, yuan2022can, shen2022mred, aaar2024} and document reasoning \citep{singh2023scirepeval, dasigi2021dataset, cohan2019structural} do not specifically isolate author-unacknowledged limitations. Recent work more directly studies limitation extraction and generation: BAGELS \citep{bagels2025} collects scholarly limitations, LIMITGEN \citep{limitgen2025} evaluates limitation identification using human and synthetic instances, and LimAgents \citep{limagents2026} uses a multi-agent, retrieval-augmented generation pipeline. These resources differ in whether they include author-stated limitations, synthetic perturbations, or external retrieval. \bench{} complements them by applying a concept-level unstated filter and preserving reviewer or citation provenance for each reference limitation.

\section{\method{} Framework}
\label{sec:framework}

As illustrated in Fig. \ref{fig:main_architecture}, \method{} systematically extracts unacknowledged paper limitations through specialized, adversarial debate.

\subsection{Problem Formulation}
\label{sec:problem}
Given a scientific paper $P$ represented by a textual context containing its title, abstract, sections, figure captions, and references, our task is to produce a set of limitation records $\mathcal{L}=\{l_1,l_2,\ldots,l_n\}$. Each record $l_i=(\text{claim}_i,\text{cat}_i,\text{sev}_i,\text{ev}_i)$ contains a limitation claim, a category $\text{cat}_i\in\{$scope, methodology, theoretical, reproducibility, fairness$\}$, a severity rating $\text{sev}_i\in\{$major, minor$\}$, and supporting evidence $\text{ev}_i$ grounded in the paper text.

\paragraph{Task scope.} We study retrieval-free, single-paper limitation identification: systems receive only the target paper's textual context. General peer-review functions such as writing assessment, novelty evaluation, and missing-citation detection are outside our scope.

\subsection{Design Principle: Specialized Critique}
\label{sec:design}

The design principle behind \method{} emerges from a simple observation (\S\ref{sec:experiments}): a generalist LLM critiquing a paper activates the same analytical patterns its authors use, clustering its outputs on methodology and scope and leaving theoretical, reproducibility, and fairness concerns largely unexplored (Appendix~\ref{app:category_motivation}). Specialization breaks this gravitational pull. By constraining each agent to a narrow, failure-mode-specific lens, the framework bypasses these salient critiques and forces the discovery of genuinely unstated concerns---ones that have minimal overlap with what a generalist baseline would surface, supporting the claim that specialization recovers fundamentally different limitations rather than a refined subset of generalist outputs (Appendix~\ref{app:complementarity}).

\subsubsection{Constructing the Personas}
To execute this, we instantiate five independent branches driven by specialized system prompts (Appendix~\ref{app:prompts}) that define their analytical scope, provide target examples, and explicitly instruct the agents to ignore stated limitations.

\begin{itemize}[leftmargin=*, topsep=0pt, itemsep=0pt]
\item \textbf{Scope Skeptic.} This persona examines whether a paper's claims generalize beyond its experimental conditions. It scrutinizes dataset coverage, domain applicability, language diversity, scale assumptions, and the gap between stated contributions and evidence. The skeptic identifies cases where authors frame results as general while evidence supports narrow conclusions, flagging unstated deployment assumptions. 
\item \textbf{Methodology Skeptic.} This persona evaluates experimental design, baselines, evaluation metrics, statistical rigor, and ablation completeness. It identifies missing controls, unfair comparisons, absent variance reporting, cherry-picked metrics, and protocols that inflate reported performance. The skeptic asks whether experimental evidence supports the paper's conclusions given these methodological choices.
\item \textbf{Theoretical Skeptic.} This persona examines formal claims, proof assumptions, complexity analyses, and logical coherence. It identifies unstated theoretical assumptions, gaps between formal guarantees and implementations, and cases where empirical results contradict theory. It applies to papers with formal contributions and abstains when theoretical content is absent.
\item \textbf{Reproducibility Skeptic.} This persona assesses whether the paper provides sufficient detail for independent replication. It examines hyperparameters, data preprocessing, computational requirements, random seeds, and code/data availability. It identifies omitted implementation choices or instances where results depend on undisclosed configurations preventing faithful reproduction.
\item \textbf{Fairness Skeptic.} This persona evaluates societal impacts, biases, and ethical implications unaddressed by the paper. It examines dataset composition for demographic biases, considers dual-use potential, identifies harmed populations, and flags cases where evaluations lack subgroup equity. It is calibrated to avoid generic concerns and ground every claim in specific paper content.
\end{itemize}

While these five personas were developed to capture a holistic view of the paper, our framework is highly extensible and allows for the seamless introduction of additional specialized skeptics as needed.

\subsection{Tree Structure and Branch Independence}
\label{sec:tree}
The five skeptic personas operate as strictly parallel, independent trees. Unlike collaborative debate trees, these branches do not exchange intermediate claims and remain isolated until the post-hoc \panel{} stage. Preventing cross-branch communication during the generation phase avoids premature convergence, ensuring that early claims do not suppress diverse perspectives. We cap tree expansion at depth 1 because deeper expansion provides only a small coverage gain while substantially reducing precision (Appendix~\ref{app:depth_ablation}).

\subsection{The Four-Stage Debate}
\label{sec:debate}
Within each tree node, a limitation candidate undergoes a four-stage structured debate (full prompt templates in Appendix~\ref{app:prompts_full}). This adversarial process eliminates superficial or poorly-grounded claims while strengthening those that survive.

\noindent\textbf{Stage 1: Argue with Evidence.} The skeptic generates a limitation claim supported by a quote from the paper. Claims without grounding evidence are disadvantaged in subsequent stages, as the moderator is unlikely to rule them valid without evidence.

\noindent\textbf{Stage 2: Advocate Response.} An advocate adopts the authors' perspective to rebut the claim, arguing it is out of scope, already addressed, or speculative. This pressure eliminates superficial or easily refuted claims to maximize precision.

\noindent\textbf{Stage 3: Revise or Withdraw.} The skeptic evaluates the rebuttal and must revise the claim or withdraw it.  Surviving claims are therefore better calibrated. Common revisions include adding unstated-constraint justifications or downgrading severity if the limitation affects secondary contributions.

\noindent\textbf{Stage 4: Moderate and Expand.} A moderator evaluates the revised claim against three expansion criteria: (a) it suggests a deeper issue worth exploring, (b) evidence supports a more specific sub-claim, and (c) expansion produces genuinely new information rather than rephrasing. If at least two criteria are met, the moderator generates a child node with a more specific follow-up claim, enabling targeted expansion into one level of more specific child claims.

\subsection{\panel{}}
\label{sec:panel}
While specialization prevents groupthink, it introduces \emph{category drift}. To resolve this, the \panel{} operates as a post-hoc reconciliation stage. Each surviving claim is re-evaluated by a single LLM call considering all five skeptic perspectives simultaneously (prompt template in Appendix~\ref{app:panel}), enabling cross-category reasoning that isolated per-tree moderators cannot perform. The panel assigns exactly one action to each claim:

\begin{itemize}[leftmargin=*, noitemsep, topsep=0pt]
\item \textbf{Endorse}: The claim is valid, correctly categorized, and substantive. It passes to the final output unchanged.
\item \textbf{Merge}: The claim spans multiple analytical categories. The panel retains one claim and records its additional applicable categories.
\item \textbf{Downgrade}: The claim is valid but its severity is inflated from the panel's broader perspective. Severity is reduced with justification.
\item \textbf{Reclassify}: The claim is valid but assigned to the wrong category due to category drift. The panel reassigns it (e.g., a theoretical skeptic's concern reclassified to scope because it addresses prose over-generalization rather than a proof gap).
\item \textbf{Reject}: The claim is invalid, already stated, or too speculative. In practice, the per-node debate filters most rejectable claims before they reach the panel.
\end{itemize}

Algorithms detailing the above pipeline are discussed in Appendix~\ref{app:algorithm}.

\begin{table}[t]
 \centering
 \begin{tabular}{lr}
 \toprule
 \textbf{Statistic} & \textbf{Value} \\
 \midrule
 Total papers & 414 \\
 \quad Gold tier & 155 \\
 \quad Silver tier & 259 \\
 Total unstated limitations & 1{,}905 \\
 Mean limitations per paper & 4.6 \\
 Median limitations per paper & 4 \\
 Source: OpenReview weaknesses & 971 \\
 Source: Citation critiques & 934 \\
 \bottomrule
 \end{tabular}
 \caption{\bench{} statistics. The benchmark contains 414 real papers with externally-sourced gold limitations actively unacknowledged by the original authors.}
 \label{tab:bench_stats}
\end{table}

\begin{table*}[!t]
\small\centering
\begin{tabular}{lccccccc}
\toprule
& \multicolumn{4}{c}{\textbf{LLM-as-Judge}} & \multicolumn{3}{c}{\textbf{Hybrid Likert Panel}} \\
\cmidrule(lr){2-5} \cmidrule(lr){6-8}
\textbf{Method} & \textbf{Cov@1} & \textbf{Cov@5} & \textbf{Cov@10} & \textbf{Precision} & \textbf{Validity} & \textbf{Specificity} & \textbf{Novelty} \\
\midrule
\multicolumn{8}{l}{\textit{Baselines}} \\
\midrule
Zero-shot LLM & 0.9 & 14.2 & 18.6 & 14.4 & 3.2 & 2.6 & 2.5 \\
DIAGPaper & 2.3 & 7.0 & 14.5 & 11.7 & 2.7 & 2.5 & 2.3 \\
Single-skeptic CoT & 8.5 & 23.7 & 32.6 & 22.5 & 3.5 & 3.4 & 3.2 \\
\midrule
\multicolumn{8}{l}{\textit{Ablations}} \\
\midrule
No-Branching & 2.6 & 7.6 & 7.6 & 24.0 & 3.7 & 3.1 & 2.6 \\
No-Expansion & 2.3 & 14.9 & 14.9 & 21.4 & 3.6 & 2.8 & 3.4 \\
No-Panel & 6.0 & 26.0 & 34.0 & 34.6 & 3.9 & 3.7 & 3.9 \\
\midrule
\textbf{ToC} & \textbf{11.5} & \textbf{27.2} & \textbf{36.1} & \textbf{40.3} & \textbf{4.3} & \textbf{4.0} & \textbf{4.2} \\
\bottomrule
\end{tabular}
\caption{Comparison of performance of \methodshort{} with baselines and ablated components. Best results in \textbf{bold}.}
\label{tab:main_results}
\end{table*}

\section{\bench{}}
\label{sec:bench}
To support rigorous evaluation of unstated limitation extraction, we introduce \bench{}. The benchmark provides externally sourced reference annotations supported by reviewer or citation evidence and filtered for acknowledgement by the original authors. General scientific-review benchmarks such as PeerRead~\citep{kang2018dataset}, NLPeer~\citep{dycke2023nlpeer}, and AAAR-1.0~\citep{aaar2024} were not designed to isolate author-unacknowledged limitations or provide typed, per-claim evidence annotations. Recent limitation-focused resources and systems include BAGELS~\citep{bagels2025}, LIMITGEN~\citep{limitgen2025}, and LimAgents~\citep{limagents2026}. \bench{} complements these works by applying a concept-level acknowledgement filter, preserving reviewer or citation provenance, and supporting retrieval-free, single-paper evaluation.

\subsection{Corpus Selection}
\label{sec:corpus_selection}

We curate \bench{} from major NLP, ML, and CV venues published between 2020 and 2025. Seed papers were discovered via the Semantic Scholar API with a citation-count threshold of $\geq$100 (relaxed to $\geq$50 for NLP venues). Complete corpus construction protocol is discussed in Appendix \ref{app:venues}. To ensure complementary temporal coverage, gold limitations are sourced from two channels,

\begin{itemize}[leftmargin=*, noitemsep, topsep=0pt]
\item \textbf{OpenReview Weaknesses (Contemporaneous):} We extract weaknesses from structured OpenReview comments using the OpenReview APIs (v1 and v2). After extraction and filtering, this source contributes 971 reference limitations. These are contemporaneous concerns raised during peer review and are generally assessable from the target paper at submission time.

\item \textbf{Citation Critiques (Retrospective):} We retrieve citing papers through the Semantic Scholar API and apply a regex cue filter (\texttt{unlike}, \texttt{fails to}, \texttt{overclaim}, etc.) to the surrounding citation contexts. After extraction and filtering, this source contributes 934 reference limitations. Some of these concerns emerge only in hindsight or downstream deployment.
\end{itemize}

The OpenReview subset represents the primary evaluation of contemporaneous, single-paper limitation discovery. The citation subset serves as a retrospective stress test, including concerns that may require external evidence unavailable to a retrieval-free system, and motivates future search-augmented evaluation. We partition the benchmark into a gold tier of 155 papers with at least two OpenReview weaknesses and a silver tier of 259 papers supported only by citation evidence. Both tiers undergo the same extraction and filtering pipeline.

\subsection{Extraction Pipeline}
\label{sec:extraction-pipeline}
A two-stage pipeline transforms the raw source text into structured limitation records.

\noindent\textbf{Stage 1: Candidate Extraction.} For each (paper, source) pair, an LLM extracts a structured limitation claim (category, severity, and evidence quote). The model is explicitly prompted to discard out-of-taxonomy claims, neutral citations, or issues that the target paper already acknowledges.

\noindent\textbf{Stage 2: Deduplication and Unstated Filter.} After clustering near-equivalent candidates, a secondary LLM compares the surviving claims against the target paper's self-critical sections (e.g., Limitations, Discussion, Future Work). Using concept-level matching, any limitation conceptually acknowledged by the authors, even if phrased differently, is removed.

After the full pipeline, we retain on average 4.6 limitations per paper (median 4, range 2--14), with a right-skewed distribution where broad-claim papers contribute the long tail. Table~\ref{tab:bench_stats} reports benchmark statistics; Appendices~\ref{app:dataset_sample} and~\ref{app:dataset_spec} provide an example record and the released JSON schema.

\noindent\textbf{Human validation.} Two independent evaluators, neither involved in system development, manually audited a stratified sample of 100 reference limitations spanning both sources and all five categories. They assessed claim fidelity, category assignment, whether each limitation was unstated, and severity calibration. Agreement was high for claim fidelity (Cohen's $\kappa=0.88$), category assignment ($\kappa=0.85$), and unstated status ($\kappa=0.82$); ordinal severity judgments achieved Spearman's $\rho=0.84$. Appendix~\ref{app:evaluators} provides the complete protocol.

\section{Experiments}
\label{sec:experiments}

\subsection{Methods Compared}
\label{sec:methods_compared}
We evaluate our framework (\textbf{ToC+Panel}) against three baselines and three ablations. To establish baseline performance, we compare against a single-shot \textbf{Zero-shot LLM}, a strong \textbf{Single-skeptic CoT} agent, and \textbf{DIAGPaper} \citep{diagpaper2024}. To isolate the impact of our core mechanisms, we systematically ablate the framework via \textbf{No-Branching}, \textbf{No-Expansion}, and \textbf{ToC no-Panel}. Complete baseline and ablation setup details are provided in Appendix~\ref{app:baselines}, prompts used by baselines in Appendix~\ref{app:baseline_prompts}, and per-method computational requirements (API calls, cost, latency) in Appendix~\ref{app:compute}.

\subsection{Evaluation Protocol}
\label{sec:eval_protocol}
We evaluate all methods on a held-out, stratified set of 100 papers out of 414 papers from \bench{}. Following recent precedents for complex evaluation tasks \citep{zheng2023judging}, we employ an LLM-as-judge protocol. For each system-generated candidate $l_s$ and gold limitation $l_g$, the judge determines whether they substantially address the same core weakness, accommodating natural variations in phrasing. Based on these matches, we compute \textbf{Coverage@K} and \textbf{Precision}.

Beyond automated matching, we conduct a qualitative hybrid  Likert evaluation to assess dimensions that binary matching misses. A panel of four independent evaluators (a PhD scholar, an undergraduate researcher, GPT-4o, and Claude Opus 4.7) rates a stratified sample of generated limitations on a 1--5 Likert scale across three criteria: \textbf{Validity}, \textbf{Specificity}, and \textbf{Novelty}. Complete metric formulations are detailed in Appendix~\ref{app:metrics} while implementation and evaluator details are provided in Appendices~\ref{app:implementation} and ~\ref{app:evaluators}.

\begin{table}[t]
\centering
\setlength{\tabcolsep}{5pt}
\begin{tabular}{lrc}
\toprule
\textbf{Panel Action} & \textbf{Count} & \textbf{\%} \\
\midrule
Endorse & 510 & 56.1 \\
Merge & 171 & 18.8 \\
Downgrade & 155 & 17.1 \\
Reclassify & 70 & 7.7 \\
Reject & 2 & 0.2 \\
\midrule
Total claims reviewed & 908 & 100.0 \\
Modification rate & 398 & 43.8 \\
\bottomrule
\end{tabular}
\caption{\panel{} verdict distribution for all panel actions.}
\label{tab:panel_verdicts}
\end{table}

\begin{table}[!t]
\centering
\setlength{\tabcolsep}{4pt}
\begin{tabular}{lccc}
\toprule
\textbf{Gold Source} & \textbf{Cov@10} & \textbf{Gold pool} & \textbf{Papers} \\
\midrule
OR weaknesses & 45.1\% & 315 & 51 \\
Citation critiques  & 31.8\% & 260 & 95 \\
\midrule
Combined      & 34.0\% & 575 & 100 \\
\bottomrule
\end{tabular}
\caption{Per-source coverage for ToC no-Panel (OR: OpenReview). Counts in the Papers column do not sum across rows because a paper can contribute gold limitations to both pools; the Combined row reports the total $100$ held-out papers.}
\label{tab:source_coverage}
\end{table}

\begin{table}[t]
\centering
\setlength{\tabcolsep}{4pt}
\begin{tabular}{lcc}
\toprule
\textbf{Method} & \textbf{Verbatim \%} & \textbf{Quotes} \\
\midrule
ToC+Panel & 61.6\% & 320/519 \\
ToC no-Panel & 64.6\% & 391/605 \\
Zero-shot LLM & 77.4\% & 531/686 \\
Single-skeptic CoT & 78.2\% & 611/781 \\
\bottomrule
\end{tabular}
\caption{Evidence grounding rates (\% of quoted evidence appearing verbatim in 
paper).}
\label{tab:grounding}
\end{table}

\begin{table}[!t]
\centering
\setlength{\tabcolsep}{6pt}
\begin{tabular}{lcc}
\toprule
\textbf{Configuration} & \textbf{Cov@10} & \textbf{$\Delta$} \\
\midrule
Full ToC+Panel       & \textbf{36.1} & --- \\
\midrule
No Methodology      & 28.5 & $-7.6$ \\
No Scope            & 30.2 & $-5.9$ \\
No Theoretical      & 32.8 & $-3.3$ \\
No Reproducibility  & 33.7 & $-2.4$ \\
No Fairness         & 34.3 & $-1.8$ \\
\bottomrule
\end{tabular}
\caption{Leave-one-persona-out ablation on the held-out evaluation set.
Coverage@10 and changes relative to the full ToC+Panel system are reported in
percentage points.}
\label{tab:lopo}
\end{table}

\begin{table*}[t]
\scriptsize
\centering
\definecolor{argueColor}{HTML}{F9EBEA}
\definecolor{respondColor}{HTML}{E8F6F3}
\definecolor{reviseColor}{HTML}{FCF3CF}
\definecolor{modColor}{HTML}{EBF5FB}
\definecolor{panelColor}{HTML}{F4ECF7}

\begin{tabular}{p{0.10\linewidth}p{0.84\linewidth}}
\toprule
\textbf{Stage} & \textbf{Transcript} \\
\midrule
\multicolumn{2}{p{0.94\linewidth}}{\textbf{Case 1.} \textit{Sharpening: Scope skeptic on DDSP.} Surface concern $\to$ specific architectural claim via one advocate--rebuttal cycle.} \\
\midrule
\rowcolor{argueColor} \textit{Argue} & The abstract claims DDSP enables ``broad applicability'' and ``high-fidelity generation,'' yet all experiments are on monophonic pitched instruments (violin, flute, trumpet) and singing voice only. \\[3pt]
\rowcolor{respondColor} \textit{Respond} & We acknowledge the concern. The paper scopes experimental contributions to audio synthesis, noting broader applicability as a direction rather than a demonstrated claim. \\[3pt]
\rowcolor{reviseColor} \textit{Revise} & Even within audio synthesis, experiments are narrow: only monophonic, pitched, single-instrument sounds. The abstract's ``high-fidelity generation'' framing is misleading given this restriction. \\[3pt]
\rowcolor{modColor} \textit{Moderate} & \textbf{Valid/major.} Expand: (1) limitations of harmonic-plus-noise architecture, (2) absence of multi-instrument or polyphonic evaluation. \\
\midrule
\multicolumn{2}{p{0.94\linewidth}}{\textbf{Case 2.} \textit{Withdrawal: Theoretical skeptic on DDSP.} Skeptic concedes when paper has no formal content; branch terminates without a claim.} \\
\midrule
\rowcolor{argueColor} \textit{Argue} & The paper does not state formal theorems, lemmas, or proofs whose internal consistency could be checked\ldots\ Claims about expressivity rely on well-known DSP results rather than novel derivations. \\[3pt]
\rowcolor{respondColor} \textit{Respond} & We agree with the reviewer's assessment\ldots\ The expressivity properties we reference are well-established in the DSP literature. \\[3pt]
\rowcolor{reviseColor} \textit{Revise} & There is no proof-level concern to raise.\ \ \emph{[concedes]} \\[3pt]
\rowcolor{modColor} \textit{Moderate} & \textbf{Deflected.} Both parties agree; branch terminates. \\
\midrule
\multicolumn{2}{p{0.94\linewidth}}{\textbf{Case 3.} \textit{Panel reclassification: Theoretical skeptic on an exploration paper.} Moderator rules valid in the original category; panel reassigns to the correct one.} \\
\midrule
\rowcolor{argueColor} \textit{Argue} & The paper claims the reward ``encourag[es] the agent to repeatedly revisit all states''\ldots\ No formal argument or proof is provided that the $k$-NN episodic bonus guarantees coverage. \\[3pt]
\rowcolor{respondColor} \textit{Respond} & We acknowledge that ``all states'' is an informal aspiration rather than a formally proven guarantee. \\[3pt]
\rowcolor{modColor} \textit{Moderate} & \textbf{Valid (theoretical).} The published text conflates a local mechanism with a global coverage property, neither proven nor validated. \\[3pt]
\rowcolor{panelColor} \textit{Panel} & \textbf{Reclassify $\to$ scope/minor.} ``The issue is imprecise natural language that overstates the method's guarantees\ldots\ the informal framing oversells generalization.'' \\
\bottomrule
\end{tabular}
\caption{Three debate transcripts illustrating the framework's core dynamics.}
\label{tab:case_debate}
\end{table*}

\subsection{Evaluation}
\label{sec:main_results}

Table~\ref{tab:main_results} presents our main results across both automated and Likert-panel evaluations.

\noindent\textbf{\method{} achieves the best performance across all metrics.} ToC+Panel reaches 36.1\% Coverage@10 and 40.3\% Precision, outperforming all baselines and ablations. The hybrid Likert panel also rates ToC+Panel highest in Validity (4.3), Specificity (4.0), and Novelty (4.2), exceeding the strongest baseline, Single-skeptic CoT, by 0.8, 0.6, and 1.0 points, respectively. A per-skeptic decomposition of the framework's outputs shows that all five personas contribute to the performance of the framework (Appendices~\ref{app:per_skeptic} and \ref{app:significance}).

\noindent\textbf{Each component contributes a distinct quality dimension.} \emph{Specialization} drives coverage and novelty; removing it (No-Branching) collapses Cov@10 to 7.6\% and Novelty to 2.6, as generalists default to acknowledged flaws. ToC's specialized branches also yield substantially broader category coverage than any baseline (Appendix~\ref{app:category}, Table~\ref{tab:category_dist}), expanding fairness representation from $<$3\% across baselines to 18.5\%. \emph{Expansion} forces specificity; without it, claims remain overly broad (Specificity 2.8, Cov@10 14.9\%). Finally, \emph{Panel review} provides calibration, intelligently merging claims to lift Precision (+5.7pp) and hybrid Likert metrics (+0.3--0.4) over ToC no-Panel.

\noindent\textbf{Baselines plateau on the obvious.} Zero-shot prompting (Validity 3.2, Specificity 2.6, Novelty 2.5) and Single-skeptic CoT (3.5, 3.4, 3.2) generate claims rated as only partially valid and largely pre-acknowledged. Their sub-3.5 Novelty scores indicate a failure to surface genuine blind spots. Although Single-skeptic CoT improves upon zero-shot via structured sequential reasoning, it lacks the adversarial debate and panel reconciliation necessary to transcend self-reported flaws, leaving gaps of 0.8 in Validity, 0.6 in Specificity, and 1.0 in Novelty behind ToC+Panel across all Likert dimensions; Appendix~\ref{app:output_comparison} shows side-by-side top-3 outputs from each method on the DDSP paper, illustrating the qualitative gap.

\noindent\textbf{Review-replication systems fail at this task.} DIAGPaper achieves the lowest scores across all automated and panel metrics (e.g., 11.7\% Precision, 2.7 Validity). Its reviewer--author dialogue generates verbose, generic feedback that overlaps with stated limitations. These uniformly low hybrid Likert ratings confirm its outputs are fundamentally weak, not merely valid concerns missing from our gold set.

\subsection{Analysis}
\label{sec:analysis}
\noindent\textbf{The panel performs substantial reconciliation.} As shown in Table~\ref{tab:panel_verdicts}, 43.8\% of valid debate nodes are modified rather than simply endorsed. The panel recalibrates severity via downgrades (17.1\%), prevents double-counting through cross-category merges (18.8\%), and corrects category drift with reclassifications (7.7\%). A near-zero rejection rate (0.2\%) confirms that earlier per-node debates successfully filter invalid claims. Thus, the panel serves as a calibrator rather than a gatekeeper, with its +0.4 Validity lift over ToC no-Panel stemming entirely from these structural adjustments.

\noindent\textbf{OpenReview-sourced limitations are 1.4$\times$ easier to recover.} Table~\ref{tab:source_coverage} shows ToC no-Panel recovers reviewer-identified weaknesses (45.1\%) substantially more often than citation critiques (31.8\%); the same asymmetry holds for Zero-shot and Single-skeptic CoT (Appendix~\ref{app:category}). This reflects a fundamental temporal gap: reviewer concerns are identifiable strictly from the paper, whereas citation critiques often rely on retrospective knowledge of subsequent developments. This inherent ceiling for single-paper systems naturally motivates retrieval-augmented extensions for future work; consistent with this, 42\% of gold limitations missed by all methods require knowledge of subsequent work (Appendix~\ref{app:errors}).

\noindent\textbf{Lower verbatim grounding reflects cross-section synthesis.} As Table~\ref{tab:grounding} shows, ToC methods generate fewer verbatim quotes (61--65\%) than baselines (77--78\%). The debate framework encourages agents to combine evidence from multiple paper sections into cohesive paraphrases, whereas single-pass methods default to quoting localized sentences suggesting that the lower verbatim rate is therefore a byproduct of deeper analytical integration rather than hallucination.

\noindent\textbf{Contribution of individual personas.} To determine whether the five-persona taxonomy contains redundant roles, we
conduct a leave-one-persona-out (LOPO) ablation on the held-out evaluation
set. In each run, we remove one specialized branch while keeping the
remaining prompts, debate stages, and inference settings unchanged.
Table~\ref{tab:lopo} shows that removing any persona reduces Coverage@10
relative to the full system. The largest reductions follow removal of the
Methodology ($-7.6$ points) and Scope ($-5.9$ points) skeptics, while the
Theoretical, Reproducibility, and Fairness skeptics contribute smaller but
measurable gains. These results indicate that all five analytical lenses
contribute to the framework's coverage.

\noindent\textbf{Qualitative Case Study.}
Table~\ref{tab:case_debate} illustrates three core mechanisms of \method{} through distinct debate transcripts (with additional examples in Appendix~\ref{app:examples}). Case~1 demonstrates \emph{adversarial sharpening}, where an advocate's pushback forces a skeptic to refine a generic concern about applicability into a highly specific architectural critique, successfully converting a surface observation into an actionable claim. Case~2 illustrates \emph{correct withdrawal}, which acts as a precision safeguard: a skeptic recognizes their theoretical lens does not apply to the paper, resulting in a deflected node (a mechanism filtering 13.6\% of all nodes to prevent manufactured flaws). Finally, Case~3 highlights \emph{panel reclassification} to resolve category drift. When an isolated theoretical moderator mislabels an overclaimed generalization as a formal proof error, the global panel catches the misalignment and reassigns it to the correct category --- a critical post-hoc correction applied to 7.7\% of valid claims that isolated moderators cannot detect.

Additional analysis using other LLMs as backbone is discussed in Appendix~\ref{app:backbone}.

\section{Conclusion}
Surfacing the unstated limitations of scientific papers is essential as the growing volume of submissions outpaces the community's reviewer capacity. We introduced \bench{}, a benchmark specifically curated for retrieval-free evaluation of author-unacknowledged limitations, containing 1{,}905 reference limitations across 414 papers sourced from OpenReview weaknesses and citation critiques, and \method{}, a multi-agent framework that deploys five category-specialized skeptic personas as parallel debate trees to extract concerns authors did not acknowledge. Our method pairs adversarial per-node debate with a cross-skeptic \panel{} stage that catches category drift and severity miscalibration before claims reach the output. With a held-out set of 100 papers evaluated by both automated metrics and human raters, we demonstrate that \method{} significantly outperforms zero-shot, single-skeptic, and review-replication baselines.

\section*{Limitations}
\label{sec:limitations}
We identify four primary limitations of this work, all reflecting the deliberate scope of \method{} rather than fixable engineering choices.

\noindent\textbf{Single-domain scope.} \bench{} draws exclusively from machine learning, natural language processing, and computer vision venues. Extending the framework to other scientific disciplines---biology, physics, the social sciences---would require domain-adapted skeptic personas (e.g., an ``experimental controls'' persona for the life sciences, a ``statistical-assumption'' persona for the social sciences) and recalibrated category taxonomies. We leave cross-disciplinary instantiation to future work.

\noindent\textbf{Text-only modality.} The framework analyzes paper prose; figures, tables, equations, and code are processed only insofar as they appear as text. Limitations that require visual reasoning over plots, formal verification of mathematical proofs, or static analysis of released code repositories fall outside our current scope. Multimodal extension via vision--language model composition is an orthogonal research direction.

\noindent\textbf{Single-paper analysis.} Each critique operates on one paper in isolation. Concerns that emerge only through cross-paper comparison---concurrent contributions, replication failures across studies, or community-level methodological drift---are not surfaced. Consistent with this, $42\%$ of gold limitations missed by all methods require knowledge of subsequent work (Appendix~\ref{app:errors}); detecting them would require retrieval-augmented multi-paper reasoning, which we leave as future work.

\noindent\textbf{Lower-bound metrics.} Gold limitations are externally grounded in reviewer critiques and follow-up citation contexts but are not exhaustive. Reported Coverage and Precision should be interpreted as conservative lower bounds rather than absolute ceilings; methods may produce additional valid concerns absent from the gold pool.

\section*{Ethics Statement}
We position \method{} as an assistive tool for human peer review, designed to highlight potential concerns rather than render automated accept/reject decisions. We recognize the dual-use risks of automated critique systems---particularly the potential for spurious objections or targeted attacks on specific authors---and advocate for responsible deployment as reviewer aids rather than autonomous evaluation mechanisms.

\bench{} is constructed from publicly accessible academic content from OpenReview and arXiv. The release contains our derived annotations, bibliographic metadata, and, only where the applicable source license permits redistribution, source text and supporting excerpts. Otherwise, it provides source identifiers, URLs, and evidence locators. Reviewer identifiers are removed. CC BY 4.0 applies only to the derived annotations and metadata created by us; all third-party content remains governed by its original license and copyright. The LLM APIs were used under their respective commercial terms.

The two human evaluators in our hybrid Likert-rating panel (Appendix~\ref{app:evaluators}) participated voluntarily, provided informed verbal consent, and rated only anonymized, method-blinded outputs after being informed that the data would evaluate an automated paper-critique system. The rating task is low-risk text annotation conducted under our organization's standard exempt-protocol guidance; benchmark construction uses publicly available text and required no separate IRB review.

AI-writing assistance (LLM-based grammar and clarity edits) was only used during manuscript preparation; all technical claims, code, experimental design, and analyses are the authors' own. The framework itself uses LLMs as both research subject and tool, as detailed throughout the paper.

\section*{Acknowledgement}
Tanmoy Chakraborty acknowledges the financial support of Anusandhan National Research Foundation (ANRF/ARG/2025/001505/QSS).

\bibliography{references}

@article{Ioannidis2007,
author = {Ioannidis, John},
year = {2007},
pages = {324--329},
title = {Limitations are not Properly Acknowledged in the Scientific Literature},
volume = {60},
journal = {Journal of clinical epidemiology},
doi = {10.1016/j.jclinepi.2006.09.011}
}

@article{Vazire2017,
    author = {Vazire, Simine},
    title = {Quality Uncertainty Erodes Trust in Science},
    journal = {Collabra: Psychology},
    volume = {3},
    number = {1},
    pages = {1},
    year = {2017},
    month = {02},
    issn = {2474-7394},
    doi = {10.1525/collabra.74},
    url = {https://doi.org/10.1525/collabra.74},
    eprint = {https://online.ucpress.edu/collabra/article-pdf/3/1/1/467141/74-765-1-pb.pdf},
}

@inproceedings{Liangetal2024,
author = {Liang, Weixin and Izzo, Zachary and Zhang, Yaohui and Lepp, Haley and Cao, Hancheng and Zhao, Xuandong and Chen, Lingjiao and Ye, Haotian and Liu, Sheng and Huang, Zhi and McFarland, Daniel A. and Zou, James Y.},
title = {Monitoring AI-modified content at scale: a case study on the impact of ChatGPT on AI conference peer reviews},
year = {2024},
publisher = {JMLR.org},
booktitle = {Proceedings of the 41st International Conference on Machine Learning},
articleno = {1192},
numpages = {46},
location = {Vienna, Austria},
series = {ICML'24}
}

@inproceedings{Cuietal2025,
    title = "Free-{MAD}: Consensus-Free Multi-Agent Debate",
    author = "Cui, Yu  and
      Fu, Hang  and
      Zhang, Haibin  and
      Wang, Licheng  and
      Zuo, Cong",
    editor = "Liakata, Maria  and
      Moreira, Viviane P.  and
      Zhang, Jiajun  and
      Jurgens, David",
    booktitle = "Findings of the {A}ssociation for {C}omputational {L}inguistics: {ACL} 2026",
    month = jul,
    year = "2026",
    address = "San Diego, California, United States",
    publisher = "Association for Computational Linguistics",
    url = "https://aclanthology.org/2026.findings-acl.1600/",
    doi = "10.18653/v1/2026.findings-acl.1600",
    pages = "31977--31997",
    ISBN = "979-8-89176-395-1",
}

@inproceedings{kargupta2025tod,
    title = "Tree-of-Debate: Multi-Persona Debate Trees Elicit Critical Thinking for Scientific Comparative Analysis",
    author = "Kargupta, Priyanka  and
      Agarwal, Ishika  and
      August, Tal  and
      Han, Jiawei",
    editor = "Che, Wanxiang  and
      Nabende, Joyce  and
      Shutova, Ekaterina  and
      Pilehvar, Mohammad Taher",
    booktitle = "Proceedings of the 63rd Annual Meeting of the Association for Computational Linguistics (Volume 1: Long Papers)",
    month = jul,
    year = "2025",
    address = "Vienna, Austria",
    publisher = "Association for Computational Linguistics",
    url = "https://aclanthology.org/2025.acl-long.1422/",
    doi = "10.18653/v1/2025.acl-long.1422",
    pages = "29378--29403",
    ISBN = "979-8-89176-251-0",
}

@inproceedings{wang2024unleashing,
    title = "Unleashing the Emergent Cognitive Synergy in Large Language Models: A Task-Solving Agent through Multi-Persona Self-Collaboration",
    author = "Wang, Zhenhailong  and
      Mao, Shaoguang  and
      Wu, Wenshan  and
      Ge, Tao  and
      Wei, Furu  and
      Ji, Heng",
    editor = "Duh, Kevin  and
      Gomez, Helena  and
      Bethard, Steven",
    booktitle = "Proceedings of the 2024 Conference of the North American Chapter of the Association for Computational Linguistics: Human Language Technologies (Volume 1: Long Papers)",
    month = jun,
    year = "2024",
    address = "Mexico City, Mexico",
    publisher = "Association for Computational Linguistics",
    url = "https://aclanthology.org/2024.naacl-long.15/",
    doi = "10.18653/v1/2024.naacl-long.15",
    pages = "257--279",
}

@inproceedings{WuIto2025,
    title = "The Hidden Strength of Disagreement: Unraveling the Consensus-Diversity Tradeoff in Adaptive Multi-Agent Systems",
    author = "Wu, Zengqing  and
      Ito, Takayuki",
    editor = "Christodoulopoulos, Christos  and
      Chakraborty, Tanmoy  and
      Rose, Carolyn  and
      Peng, Violet",
    booktitle = "Proceedings of the 2025 Conference on Empirical Methods in Natural Language Processing",
    month = nov,
    year = "2025",
    address = "Suzhou, China",
    publisher = "Association for Computational Linguistics",
    url = "https://aclanthology.org/2025.emnlp-main.772/",
    doi = "10.18653/v1/2025.emnlp-main.772",
    pages = "15277--15297",
    ISBN = "979-8-89176-332-6",
}

@article{diagpaper2024,
  title={DIAGPaper: Diagnosing Valid and Specific Weaknesses in Scientific Papers via Multi-Agent Reasoning},
  author={Zou, Zhuoyang and Ansari, Abolfazl and Zhang, Delvin Ce and Lee, Dongwon and Yin, Wenpeng},
  journal={arXiv preprint arXiv:2601.07611},
  year={2026}
}

@article{damico2024marg,
  title={Marg: Multi-agent review generation for scientific papers},
  author={D'Arcy, Mike and Hope, Tom and Birnbaum, Larry and Downey, Doug},
  journal={arXiv preprint arXiv:2401.04259},
  year={2024}
}

@inproceedings{jin2024agentreview,
    title = "{A}gent{R}eview: Exploring Peer Review Dynamics with {LLM} Agents",
    author = "Jin, Yiqiao  and
      Zhao, Qinlin  and
      Wang, Yiyang  and
      Chen, Hao  and
      Zhu, Kaijie  and
      Xiao, Yijia  and
      Wang, Jindong",
    editor = "Al-Onaizan, Yaser  and
      Bansal, Mohit  and
      Chen, Yun-Nung",
    booktitle = "Proceedings of the 2024 Conference on Empirical Methods in Natural Language Processing",
    month = nov,
    year = "2024",
    address = "Miami, Florida, USA",
    publisher = "Association for Computational Linguistics",
    url = "https://aclanthology.org/2024.emnlp-main.70/",
    doi = "10.18653/v1/2024.emnlp-main.70",
    pages = "1208--1226",
}

@inproceedings{kang2018dataset,
    title = "A Dataset of Peer Reviews ({P}eer{R}ead): Collection, Insights and {NLP} Applications",
    author = "Kang, Dongyeop  and
      Ammar, Waleed  and
      Dalvi, Bhavana  and
      van Zuylen, Madeleine  and
      Kohlmeier, Sebastian  and
      Hovy, Eduard  and
      Schwartz, Roy",
    editor = "Walker, Marilyn  and
      Ji, Heng  and
      Stent, Amanda",
    booktitle = "Proceedings of the 2018 Conference of the North {A}merican Chapter of the Association for Computational Linguistics: Human Language Technologies, Volume 1 (Long Papers)",
    month = jun,
    year = "2018",
    address = "New Orleans, Louisiana",
    publisher = "Association for Computational Linguistics",
    url = "https://aclanthology.org/N18-1149/",
    doi = "10.18653/v1/N18-1149",
    pages = "1647--1661",
}

@inproceedings{dycke2023nlpeer,
    title = "{NLP}eer: A Unified Resource for the Computational Study of Peer Review",
    author = "Dycke, Nils  and
      Kuznetsov, Ilia  and
      Gurevych, Iryna",
    editor = "Rogers, Anna  and
      Boyd-Graber, Jordan  and
      Okazaki, Naoaki",
    booktitle = "Proceedings of the 61st Annual Meeting of the Association for Computational Linguistics (Volume 1: Long Papers)",
    month = jul,
    year = "2023",
    address = "Toronto, Canada",
    publisher = "Association for Computational Linguistics",
    url = "https://aclanthology.org/2023.acl-long.277/",
    doi = "10.18653/v1/2023.acl-long.277",
    pages = "5049--5073",
}

@inproceedings{aaar2024,
author = {Lou, Renze and Xu, Hanzi and Wang, Sijia and Du, Jiangshu and Kamoi, Ryo and Lu, Xiaoxin and Xie, Jian and Sun, Yuxuan and Zhang, Yusen and Ahn, Jihyun Janice and Fang, Hongchao and Zou, Zhuoyang and Ma, Wenchao and Li, Xi and Zhang, Kai and Xia, Congying and Huang, Lifu and Yin, Wenpeng},
title = {AAAR-1.0: assessing AI's potential to assist research},
year = {2025},
publisher = {JMLR.org},
booktitle = {Proceedings of the 42nd International Conference on Machine Learning},
articleno = {1623},
numpages = {23},
location = {Vancouver, Canada},
series = {ICML'25}
}

@inproceedings{du2023improving,
author = {Du, Yilun and Li, Shuang and Torralba, Antonio and Tenenbaum, Joshua B. and Mordatch, Igor},
title = {Improving factuality and reasoning in language models through multiagent debate},
year = {2024},
publisher = {JMLR.org},
booktitle = {Proceedings of the 41st International Conference on Machine Learning},
articleno = {467},
numpages = {31},
location = {Vienna, Austria},
series = {ICML'24}
}

@ARTICLE{zhuge2024language,
  author={Zhuge, Mingchen and Liu, Haozhe and Faccio, Francesco and Ashley, Dylan R. and Csordás, Róbert and Gopalakrishnan, Anand and Hamdi, Abdullah and Hammoud, Hasan Abed Al Kader and Herrmann, Vincent and Irie, Kazuki and Kirsch, Louis and Li, Bing and Li, Guohao and Liu, Shuming and Mai, Jinjie and Pi{\k{e}}kos, Piotr and Ramesh, Aditya A. and Schlag, Imanol and Shi, Weimin and Stanić, Aleksandar and Wang, Wenyi and Wang, Yuhui and Xu, Mengmeng and Fan, Deng-Ping and Ghanem, Bernard and Schmidhuber, Jürgen},
  journal={Computational Visual Media}, 
  title={Mindstorms in Natural Language-Based Societies of Mind}, 
  year={2025},
  volume={11},
  number={1},
  pages={29-81},
  doi={10.26599/CVM.2025.9450460}
}

@article{irving2018ai,
  title={AI safety via debate},
  author={Irving, Geoffrey and Christiano, Paul and Amodei, Dario},
  journal={arXiv preprint arXiv:1805.00899},
  year={2018}
}

@inproceedings{chan2023chateval,
  title={Chateval: Towards better llm-based evaluators through multi-agent debate},
  author={Chan, Chi-Min and Chen, Weize and Su, Yusheng and Yu, Jianxuan and Xue, Wei and Zhang, Shanghang and Fu, Jie and Liu, Zhiyuan},
  booktitle={International conference on learning representations},
  volume={2024},
  pages={9079--9093},
  year={2024}
}

@inproceedings{liang2023encouraging,
    title = "Encouraging Divergent Thinking in Large Language Models through Multi-Agent Debate",
    author = "Liang, Tian  and
      He, Zhiwei  and
      Jiao, Wenxiang  and
      Wang, Xing  and
      Wang, Yan  and
      Wang, Rui  and
      Yang, Yujiu  and
      Shi, Shuming  and
      Tu, Zhaopeng",
    editor = "Al-Onaizan, Yaser  and
      Bansal, Mohit  and
      Chen, Yun-Nung",
    booktitle = "Proceedings of the 2024 Conference on Empirical Methods in Natural Language Processing",
    month = nov,
    year = "2024",
    address = "Miami, Florida, USA",
    publisher = "Association for Computational Linguistics",
    url = "https://aclanthology.org/2024.emnlp-main.992/",
    doi = "10.18653/v1/2024.emnlp-main.992",
    pages = "17889--17904",
}

@article{tyser2024reviewergpt,
  title={Reviewergpt? an exploratory study on using large language models for paper reviewing},
  author={Liu, Ryan and Shah, Nihar B},
  journal={arXiv preprint arXiv:2306.00622},
  year={2023}
}

@inproceedings{
  engel2020ddsp,
  title={DDSP: Differentiable Digital Signal Processing},
    author={Jesse Engel and Lamtharn (Hanoi) Hantrakul and Chenjie Gu and Adam Roberts},
    booktitle={International Conference on Learning Representations},
    year={2020},
    url={https://openreview.net/forum?id=B1x1ma4tDr}
}

@article{yuan2022can,
  title={Can we automate scientific reviewing?},
  author={Yuan, Weizhe and Liu, Pengfei and Neubig, Graham},
  journal={Journal of Artificial Intelligence Research},
  volume={75},
  year={2022},
  pages={171--212},
  url={https://doi.org/10.1613/jair.1.12862}
}

@inproceedings{singh2023scirepeval,
    title = "{S}ci{R}ep{E}val: A Multi-Format Benchmark for Scientific Document Representations",
    author = "Singh, Amanpreet  and
      D{'}Arcy, Mike  and
      Cohan, Arman  and
      Downey, Doug  and
      Feldman, Sergey",
    editor = "Bouamor, Houda  and
      Pino, Juan  and
      Bali, Kalika",
    booktitle = "Proceedings of the 2023 Conference on Empirical Methods in Natural Language Processing",
    month = dec,
    year = "2023",
    address = "Singapore",
    publisher = "Association for Computational Linguistics",
    url = "https://aclanthology.org/2023.emnlp-main.338/",
    doi = "10.18653/v1/2023.emnlp-main.338",
    pages = "5548--5566",
}

@article{huang2025survey,
  title={A survey on hallucination in large language models: Principles, taxonomy, challenges, and open questions},
  author={Huang, Lei and Yu, Weijiang and Ma, Weitao and Zhong, Weihong and Feng, Zhangyin and Wang, Haotian and Chen, Qianglong and Peng, Weihua and Feng, Xiaocheng and Qin, Bing and others},
  journal={ACM Transactions on Information Systems},
  volume={43},
  number={2},
  pages={1--55},
  year={2025},
  publisher={ACM New York, NY},
  url={https://doi.org/10.1145/3703155}
}

@inproceedings{zheng2023judging,
 author = {Zheng, Lianmin and Chiang, Wei-Lin and Sheng, Ying and Zhuang, Siyuan and Wu, Zhanghao and Zhuang, Yonghao and Lin, Zi and Li, Zhuohan and Li, Dacheng and Xing, Eric and Zhang, Hao and Gonzalez, Joseph and Stoica, Ion},
 booktitle = {Advances in Neural Information Processing Systems},
 doi = {10.52202/075280-2020},
 editor = {A. Oh and T. Naumann and A. Globerson and K. Saenko and M. Hardt and S. Levine},
 pages = {46595--46623},
 publisher = {Curran Associates, Inc.},
 title = {Judging LLM-as-a-Judge with MT-Bench and Chatbot Arena},
 url = {https://proceedings.neurips.cc/paper_files/paper/2023/file/91f18a1287b398d378ef22505bf41832-Paper-Datasets_and_Benchmarks.pdf},
 volume = {36},
 year = {2023}
}

@article{Kapoor2024,
author = {Sayash Kapoor  and Emily M. Cantrell  and Kenny Peng  and Thanh Hien Pham  and Christopher A. Bail  and Odd Erik Gundersen  and Jake M. Hofman  and Jessica Hullman  and Michael A. Lones  and Momin M. Malik  and Priyanka Nanayakkara  and Russell A. Poldrack  and Inioluwa Deborah Raji  and Michael Roberts  and Matthew J. Salganik  and Marta Serra-Garcia  and Brandon M. Stewart  and Gilles Vandewiele  and Arvind Narayanan },
title = {REFORMS: Consensus-based Recommendations for Machine-learning-based Science},
journal = {Science Advances},
volume = {10},
number = {18},
pages = {eadk3452},
year = {2024},
doi = {10.1126/sciadv.adk3452},
URL = {https://www.science.org/doi/abs/10.1126/sciadv.adk3452},
eprint = {https://www.science.org/doi/pdf/10.1126/sciadv.adk3452},
}

@article{semmelrock2025reproducibility,
author = {Semmelrock, Harald and Ross-Hellauer, Tony and Kopeinik, Simone and Theiler, Dieter and Haberl, Armin and Thalmann, Stefan and Kowald, Dominik},
title = {Reproducibility in machine-learning-based research: Overview, barriers, and drivers},
journal = {AI Magazine},
volume = {46},
number = {2},
pages = {e70002},
doi = {https://doi.org/10.1002/aaai.70002},
url = {https://onlinelibrary.wiley.com/doi/abs/10.1002/aaai.70002},
eprint = {https://onlinelibrary.wiley.com/doi/pdf/10.1002/aaai.70002},
year = {2025}
}

@inproceedings{
Wengetal2025,
title={Do as We Do, Not as You Think: the Conformity of Large Language Models},
author={Zhiyuan Weng and Guikun Chen and Wenguan Wang},
booktitle={The Thirteenth International Conference on Learning Representations},
year={2025},
url={https://openreview.net/forum?id=st77ShxP1K}
}

@inproceedings{Choietal2025,
    title = "An Empirical Study of Group Conformity in Multi-Agent Systems",
    author = "Choi, Min  and
      Kim, Keonwoo  and
      Chae, Sungwon  and
      Baek, Sangyeop",
    editor = "Che, Wanxiang  and
      Nabende, Joyce  and
      Shutova, Ekaterina  and
      Pilehvar, Mohammad Taher",
    booktitle = "Findings of the Association for Computational Linguistics: ACL 2025",
    month = jul,
    year = "2025",
    address = "Vienna, Austria",
    publisher = "Association for Computational Linguistics",
    url = "https://aclanthology.org/2025.findings-acl.265/",
    doi = "10.18653/v1/2025.findings-acl.265",
    pages = "5123--5139",
    ISBN = "979-8-89176-256-5",
}

@inproceedings{wang2023large,
    title = "Large Language Models are not Fair Evaluators",
    author = "Wang, Peiyi  and
      Li, Lei  and
      Chen, Liang  and
      Cai, Zefan  and
      Zhu, Dawei  and
      Lin, Binghuai  and
      Cao, Yunbo  and
      Kong, Lingpeng  and
      Liu, Qi  and
      Liu, Tianyu  and
      Sui, Zhifang",
    editor = "Ku, Lun-Wei  and
      Martins, Andre  and
      Srikumar, Vivek",
    booktitle = "Proceedings of the 62nd Annual Meeting of the Association for Computational Linguistics (Volume 1: Long Papers)",
    month = aug,
    year = "2024",
    address = "Bangkok, Thailand",
    publisher = "Association for Computational Linguistics",
    url = "https://aclanthology.org/2024.acl-long.511/",
    doi = "10.18653/v1/2024.acl-long.511",
    pages = "9440--9450",
}

@inproceedings{shen2022mred,
    title = "{MR}e{D}: A Meta-Review Dataset for Structure-Controllable Text Generation",
    author = "Shen, Chenhui  and
      Cheng, Liying  and
      Zhou, Ran  and
      Bing, Lidong  and
      You, Yang  and
      Si, Luo",
    editor = "Muresan, Smaranda  and
      Nakov, Preslav  and
      Villavicencio, Aline",
    booktitle = "Findings of the Association for Computational Linguistics: ACL 2022",
    month = may,
    year = "2022",
    address = "Dublin, Ireland",
    publisher = "Association for Computational Linguistics",
    url = "https://aclanthology.org/2022.findings-acl.198/",
    doi = "10.18653/v1/2022.findings-acl.198",
    pages = "2521--2535",
}

@inproceedings{dasigi2021dataset,
    title = "A Dataset of Information-Seeking Questions and Answers Anchored in Research Papers",
    author = "Dasigi, Pradeep  and
      Lo, Kyle  and
      Beltagy, Iz  and
      Cohan, Arman  and
      Smith, Noah A.  and
      Gardner, Matt",
    editor = "Toutanova, Kristina  and
      Rumshisky, Anna  and
      Zettlemoyer, Luke  and
      Hakkani-Tur, Dilek  and
      Beltagy, Iz  and
      Bethard, Steven  and
      Cotterell, Ryan  and
      Chakraborty, Tanmoy  and
      Zhou, Yichao",
    booktitle = "Proceedings of the 2021 Conference of the North American Chapter of the Association for Computational Linguistics: Human Language Technologies",
    month = jun,
    year = "2021",
    address = "Online",
    publisher = "Association for Computational Linguistics",
    url = "https://aclanthology.org/2021.naacl-main.365/",
    doi = "10.18653/v1/2021.naacl-main.365",
    pages = "4599--4610",
}

@inproceedings{cohan2019structural,
    title = "Structural Scaffolds for Citation Intent Classification in Scientific Publications",
    author = "Cohan, Arman  and
      Ammar, Waleed  and
      van Zuylen, Madeleine  and
      Cady, Field",
    editor = "Burstein, Jill  and
      Doran, Christy  and
      Solorio, Thamar",
    booktitle = "Proceedings of the 2019 Conference of the North {A}merican Chapter of the Association for Computational Linguistics: Human Language Technologies, Volume 1 (Long and Short Papers)",
    month = jun,
    year = "2019",
    address = "Minneapolis, Minnesota",
    publisher = "Association for Computational Linguistics",
    url = "https://aclanthology.org/N19-1361/",
    doi = "10.18653/v1/N19-1361",
    pages = "3586--3596",
}

@article{landis1977measurement,
 ISSN = {0006341X, 15410420},
 URL = {http://www.jstor.org/stable/2529310},
 author = {J. Richard Landis and Gary G. Koch},
 journal = {Biometrics},
 number = {1},
 pages = {159--174},
 publisher = {International Biometric Society},
 title = {The Measurement of Observer Agreement for Categorical Data},
 urldate = {2026-05-25},
 volume = {33},
 year = {1977}
}

@article{mishra2024flame,
author = {Mishra, Sahil and Sudev, Ujjwal and Chakraborty, Tanmoy},
title = {FLAME: Self-Supervised Low-Resource Taxonomy Expansion Using Large Language Models},
year = {2026},
issue_date = {December 2026},
publisher = {Association for Computing Machinery},
address = {New York, NY, USA},
volume = {17},
number = {6},
issn = {2157-6904},
url = {https://doi.org/10.1145/3709007},
doi = {10.1145/3709007},
journal = {ACM Trans. Intell. Syst. Technol.},
month = aug,
articleno = {132},
numpages = {25}
}

@inproceedings{mishra2025lorex,
    title = "Rank, Chunk and Expand: Lineage-Oriented Reasoning for Taxonomy Expansion",
    author = "Mishra, Sahil  and
      Arjun, Kumar  and
      Chakraborty, Tanmoy",
    editor = "Che, Wanxiang  and
      Nabende, Joyce  and
      Shutova, Ekaterina  and
      Pilehvar, Mohammad Taher",
    booktitle = "Findings of the Association for Computational Linguistics: ACL 2025",
    month = jul,
    year = "2025",
    address = "Vienna, Austria",
    publisher = "Association for Computational Linguistics",
    url = "https://aclanthology.org/2025.findings-acl.671/",
    doi = "10.18653/v1/2025.findings-acl.671",
    pages = "12935--12953",
    ISBN = "979-8-89176-256-5",
}

@inproceedings{mishra2026quantaxo,
  title={Quantaxo: A quantum approach to self-supervised taxonomy expansion},
  author={Mishra, Sahil and Patni, Avi and Chatterjee, Niladri and Chakraborty, Tanmoy},
  booktitle={Proceedings of the AAAI Conference on Artificial Intelligence},
  volume={40},
  number={38},
  pages={32501--32509},
  year={2026}
}

@inproceedings{mishra2026taxobell,
author = {Mishra, Sahil and Srinivasan, Srinitish and Bedathur, Srikanta and Chakraborty, Tanmoy},
title = {TaxoBell: Gaussian Box Embeddings for Self-Supervised Taxonomy Expansion},
year = {2026},
isbn = {9798400723070},
publisher = {Association for Computing Machinery},
address = {New York, NY, USA},
url = {https://doi.org/10.1145/3774904.3792674},
doi = {10.1145/3774904.3792674},
booktitle = {Proceedings of the ACM Web Conference 2026},
pages = {4326--4337},
numpages = {12},
location = {United Arab Emirates},
series = {WWW '26}
}

@inproceedings{
mishra2026polaris,
title={Polaris: Coupled Orbital Polar Embeddings for Hierarchical Concept Learning},
author={Sahil Mishra and Srinitish Srinivasan and Sourish Dasgupta and Tanmoy Chakraborty},
booktitle={Forty-third International Conference on Machine Learning},
year={2026},
url={https://openreview.net/forum?id=FQNYtjPBxv}
}

@inproceedings{bagels2025,
    title = "{BAGELS}: Benchmarking the Automated Generation and Extraction of Limitations from Scholarly Text",
    author = "Al Azher, Ibrahim  and
      Mokarrama, Miftahul Jannat  and
      Guo, Zhishuai  and
      Choudhury, Sagnik Ray  and
      Alhoori, Hamed",
    editor = "Christodoulopoulos, Christos  and
      Chakraborty, Tanmoy  and
      Rose, Carolyn  and
      Peng, Violet",
    booktitle = "Findings of the Association for Computational Linguistics: EMNLP 2025",
    month = nov,
    year = "2025",
    address = "Suzhou, China",
    publisher = "Association for Computational Linguistics",
    url = "https://aclanthology.org/2025.findings-emnlp.1050/",
    doi = "10.18653/v1/2025.findings-emnlp.1050",
    pages = "19279--19294",
    ISBN = "979-8-89176-335-7",
}

@article{limagents2026,
  title={Multi-Agent LLMs for Generating Research Limitations},
  author={Azher, Ibrahim Al and Guo, Zhishuai and Alhoori, Hamed},
  journal={arXiv preprint arXiv:2601.11578},
  year={2026}
}

@inproceedings{limitgen2025,
    title = "Can {LLM}s Identify Critical Limitations within Scientific Research? A Systematic Evaluation on {AI} Research Papers",
    author = "Xu, Zhijian  and
      Zhao, Yilun  and
      Patwardhan, Manasi  and
      Vig, Lovekesh  and
      Cohan, Arman",
    editor = "Che, Wanxiang  and
      Nabende, Joyce  and
      Shutova, Ekaterina  and
      Pilehvar, Mohammad Taher",
    booktitle = "Proceedings of the 63rd Annual Meeting of the Association for Computational Linguistics (Volume 1: Long Papers)",
    month = jul,
    year = "2025",
    address = "Vienna, Austria",
    publisher = "Association for Computational Linguistics",
    url = "https://aclanthology.org/2025.acl-long.1009/",
    doi = "10.18653/v1/2025.acl-long.1009",
    pages = "20652--20706",
    ISBN = "979-8-89176-251-0",
}

@article{wynn2025,
  title={Talk isn't always cheap: Understanding failure modes in multi-agent debate},
  author={Wynn, Andrea and Satija, Harsh and Hadfield, Gillian},
  journal={arXiv preprint arXiv:2509.05396},
  year={2025}
}

\appendix

\section*{\centering Appendix}

\section{Empirical Motivation for Specialized Personas}
\label{app:category_motivation}

Following the design principle introduced in \S\ref{sec:design}, this appendix provides empirical evidence for the claim that generalist LLMs gravitate toward the methodology+scope concentration of stated reviewer feedback, leaving theoretical, reproducibility, and fairness dimensions under-explored. Table~\ref{tab:category_concentration} confirms this empirically on our held-out evaluation set. All three generalist baselines concentrate 64.5--88.8\% of their outputs in the scope and methodology categories, closely mirroring the gold distribution's 77.4\% concentration in those two categories---which is itself a property of the reviewer-and-citation sources used to construct the benchmark, not of the underlying space of real limitations. \method{}'s specialized debate trees substantially break this pattern: ToC+Panel produces only 50.8\% of outputs in methodology+scope and substantially expands fairness coverage from near-zero (0.6--2.7\% across baselines) to 18.5\%, with parallel expansion in reproducibility (4.8--13.0\% $\to$ 18.5\%).

The KL divergence of each method's category distribution from the gold distribution makes this concrete. Zero-shot ($0.013$), Single-skeptic CoT ($0.053$), and DIAGPaper ($0.084$) all match the gold closely because they inherit the same reviewer attention bias the gold reflects. ToC+Panel diverges substantially ($0.492$) precisely because it surfaces concerns the reviewer-and-citation sources systematically miss. We interpret this as a feature rather than a weakness: the framework's over-representation of fairness and reproducibility relative to the gold reflects a calibrated correction of an upstream sourcing bias rather than hallucinated content. Our hybrid Likert panel supports this interpretation, with ToC+Panel's outputs receiving the highest Validity, Specificity, and Novelty ratings of any method (4.3, 4.0, 4.2 on the 1--5 Likert scale).

\begin{table}[b]
\scriptsize\centering
\setlength{\tabcolsep}{3pt}
\begin{tabular}{lrrrrrr}
\toprule
\textbf{Source / Method} & \textbf{Sc.} & \textbf{Me.} & \textbf{Th.} & \textbf{Re.} & \textbf{Fa.} & \textbf{KL} \\
\midrule
Gold ($n{=}575$)     & 40.9\% & 36.5\% & 14.8\% &  7.0\% &  0.9\% & --- \\
\midrule
Zero-shot LLM      & 35.6\% & 36.9\% & 16.2\% & 10.6\% &  0.6\% & 0.013 \\
Single-skeptic CoT     & 29.0\% & 35.5\% & 20.5\% & 13.0\% &  2.0\% & 0.053 \\
DIAGPaper (reimpl.)    & 46.5\% & 42.3\% &  3.6\% &  4.8\% &  2.7\% & 0.084 \\
\textbf{ToC+Panel (ours)}  & \textbf{27.4\%} & \textbf{23.4\%} & \textbf{12.1\%} & \textbf{18.5\%} & \textbf{18.5\%} & \textbf{0.492} \\
\bottomrule
\end{tabular}
\caption{Category distribution (\%) of outputs per method, alongside the gold distribution from the 100-paper held-out set. Sc.=scope, Me.=methodology, Th.=theoretical, Re.=reproducibility, Fa.=fairness.}
\label{tab:category_concentration}
\end{table}

\section{Skeptic Persona Priors}
\label{app:prompts}

Following the persona-specialization design introduced in \S\ref{sec:design}, each skeptic receives a category-specific system-prompt prior prepended to the shared debate instructions. The five priors are:

\begin{tcolorbox}[colback=blue!3, colframe=blue!40, title=\textbf{Scope Skeptic Prior}, fonttitle=\bfseries\small, fontupper=\small\ttfamily, arc=2mm, boxrule=0.5pt, left=2mm, right=2mm, top=2mm, bottom=2mm]
You are a Scope Skeptic. You look for mismatches between the paper's \textit{claims} and what it \textit{actually tested}.\\
Probe for:\\
- Claims of generality supported only by single-domain benchmarks\\
- Claims of scaling supported only at smaller sizes\\
- Cross-lingual/cross-task claims tested only on English/one task\\
- ``Zero-shot'' claims relying on unaudited pretraining data\\
- Implicit distribution assumptions mismatched to deployment
\end{tcolorbox}

\begin{tcolorbox}[colback=green!3, colframe=green!40, title=\textbf{Methodology Skeptic Prior}, fonttitle=\bfseries\small, fontupper=\small\ttfamily, arc=2mm, boxrule=0.5pt, left=2mm, right=2mm, top=2mm, bottom=2mm]
You are a Methodology Skeptic. You look for holes in the experimental setup that bias the comparison.\\
Probe for:\\
- Baselines tuned less carefully than the proposed method\\
- Missing obvious baselines\\
- Evaluation metrics favoring the method by construction\\
- Data leakage: test overlap with train or pretraining\\
- Ablations that add instead of remove\\
- Single-dataset claims where multi-dataset is standard
\end{tcolorbox}

\begin{tcolorbox}[colback=purple!3, colframe=purple!40, title=\textbf{Theoretical Skeptic Prior}, fonttitle=\bfseries\small, fontupper=\small\ttfamily, arc=2mm, boxrule=0.5pt, left=2mm, right=2mm, top=2mm, bottom=2mm]
You are a Theoretical Skeptic. You check soundness of proofs, definitions, assumptions.\\
Probe for:\\
- Assumptions in theorems that fail in the experimental setting\\
- Definitions drifting between formal and experimental sections\\
- Circular reasoning (metric designed to reward the method)\\
- Proofs skipping non-obvious steps\\
- Reductions to prior work that change the setting
\end{tcolorbox}

\begin{tcolorbox}[colback=orange!3, colframe=orange!40, title=\textbf{Reproducibility Skeptic Prior}, fonttitle=\bfseries\small, fontupper=\small\ttfamily, arc=2mm, boxrule=0.5pt, left=2mm, right=2mm, top=2mm, bottom=2mm]
You are a Reproducibility Skeptic. You assess whether the paper can be faithfully reproduced.\\
Probe for:\\
- Missing/vague hyperparameters (LR, batch size, seed, schedule)\\
- Single-seed results where variance matters\\
- Undisclosed data preprocessing\\
- Unavailable/undocumented training data\\
- Hardware/random-init sensitivity not characterized\\
- Evaluation code not released
\end{tcolorbox}

\begin{tcolorbox}[colback=red!3, colframe=red!40, title=\textbf{Fairness Skeptic Prior}, fonttitle=\bfseries\small, fontupper=\small\ttfamily, arc=2mm, boxrule=0.5pt, left=2mm, right=2mm, top=2mm, bottom=2mm]
You are a Fairness Skeptic. You look for subgroup gaps and demographic blind spots.\\
Probe for:\\
- Aggregate metrics hiding disparities across race/gender/age/language\\
- Training data with known demographic skew, no audit\\
- Benchmarks underrepresenting minority groups\\
- Claims of fairness based on a single metric\\
- Downstream applications with disparate-impact risk not discussed
\end{tcolorbox}

Each prior is followed by the shared instruction block: ``\texttt{You are rigorous but fair. You only raise a concern if you can ground it in the paper text. Never manufacture a quote. Be specific and terse. If you cannot find a concrete concern, return a very short description and leave the evidence quote empty; the moderator will terminate this branch.}''

\section{Full Prompt Templates}
\label{app:prompts_full}

Following the four-stage debate procedure introduced in \S\ref{sec:debate} and the \panel{} reconciliation in \S\ref{sec:panel}, this appendix lists the complete prompt templates for each stage of the \method{} debate pipeline and the evaluation judge. All templates are transcribed directly from the codebase; field placeholders are shown as \verb|{field_name}|.

\subsection{Stage 1: Skeptic Argue}

The skeptic receives its category-specific prior (shown in \S\ref{app:prompts}) plus the paper context, and is asked to propose one evidence-grounded concern. Temperature: 0.3.

\begin{tcolorbox}[colback=gray!5, colframe=gray!60, title=\textbf{Skeptic Argue}, fonttitle=\bfseries\small, fontupper=\small\ttfamily, arc=2mm, boxrule=0.5pt, left=2mm, right=2mm, top=2mm, bottom=2mm]
\textbf{System:}\\
\{skeptic\_prior\}\\
You are rigorous but fair. You only raise a concern if you can ground it in the paper text. Never manufacture a quote. If the paper contradicts your concern, say so. Be specific and terse.\\[0.3em]
Paper context: \{paper\_context\}

\vspace{0.5em}
\textbf{User:}\\
The topic of this debate node is: "\{topic\}".\\
Topic description: \{topic\_description\}\\[0.3em]
Propose ONE specific concern under this topic. Ground it in a specific quote from the paper. If you cannot find a concrete concern, return a very short description and leave evidence\_quote empty; the moderator will terminate this branch.

\vspace{0.5em}
\textbf{Output:}\\
\{"topic": "...", "description": "...", "evidence\_quote": "...", "evidence\_section": "...", "severity\_guess": "minor|major"\}
\end{tcolorbox}

\subsection{Stage 2: Paper Advocate Response}

The advocate adopts the authors' perspective and is given the full paper context. It must ground its rebuttal in actual paper content and is explicitly instructed not to fabricate. Temperature: 0.2.

\begin{tcolorbox}[colback=gray!5, colframe=gray!60, title=\textbf{Paper Advocate Response}, fonttitle=\bfseries\small, fontupper=\small\ttfamily, arc=2mm, boxrule=0.5pt, left=2mm, right=2mm, top=2mm, bottom=2mm]
\textbf{System:}\\
You are the authors of the paper below. You will be asked to respond to a critical concern raised by a skeptical reviewer.\\[0.3em]
Rules:\\
- You MUST ground your response in the actual content of the paper, quoting where possible.\\
- If the paper genuinely does not address the concern, acknowledge the limitation honestly. Do not fabricate.\\
- Be specific and terse. Cite section names where relevant.\\[0.3em]
PAPER: \{paper\_context\}

\vspace{0.5em}
\textbf{User:}\\
A skeptical reviewer has raised a concern:\\
~~Concern topic: "\{topic\}"\\
~~Description: \{description\}\\
~~Evidence quoted: "\{evidence\_quote\}" (section: \{evidence\_section\})\\[0.3em]
Respond. You may acknowledge, rebut with a counter-quote, or clarify.

\vspace{0.5em}
\textbf{Output:}\\
\{"acknowledges": true|false, "response": "...", "citation\_quote": "..."\}
\end{tcolorbox}

\subsection{Stage 3: Skeptic Revise or Withdraw}

After seeing the advocate's rebuttal, the skeptic must sharpen or concede. A \texttt{concedes=true} output terminates the node immediately. Temperature: 0.2.

\begin{tcolorbox}[colback=gray!5, colframe=gray!60, title=\textbf{Skeptic Revise/Withdraw}, fonttitle=\bfseries\small, fontupper=\small\ttfamily, arc=2mm, boxrule=0.5pt, left=2mm, right=2mm, top=2mm, bottom=2mm]
\textbf{System:}\\
\{skeptic\_prior + paper\_context\} (same as Stage 1)

\vspace{0.5em}
\textbf{User:}\\
You raised: \{original\_description\}\\
Evidence quoted: "\{evidence\_quote\}"\\[0.3em]
The paper responded:\\
~~response: \{paper\_response\}\\
~~acknowledges: \{acknowledges\}\\
~~counter-quote: "\{citation\_quote\}"\\[0.3em]
Revise your concern. If the paper's response genuinely addresses it, concede. Otherwise, sharpen focusing on what remains unaddressed.

\vspace{0.5em}
\textbf{Output:}\\
\{"revised\_description": "...", "concedes": true|false\}
\end{tcolorbox}

\subsection{Stage 4: Moderator Verdict}

The moderator sees the full debate transcript (argue, respond, revise) and rules on validity, severity, and whether to spawn child nodes. Temperature: 0.1 for consistent adjudication.

\begin{tcolorbox}[colback=gray!5, colframe=gray!60, title=\textbf{Moderator Verdict}, fonttitle=\bfseries\small, fontupper=\small\ttfamily, arc=2mm, boxrule=0.5pt, left=2mm, right=2mm, top=2mm, bottom=2mm]
\textbf{System:}\\
You are a neutral moderator judging a debate between a skeptic and a paper's authors.\\
Rule valid only if the concern is concrete, grounded in the paper, and the paper did not adequately address it.\\
Rule deflected if the paper credibly addresses the concern.\\
Rule unclear if there is not enough information.

\vspace{0.5em}
\textbf{User:}\\
Topic: \{topic\}~~Category: \{category\}\\[0.3em]
Skeptic initial concern: \{argument\_description\}\\
Evidence quoted: "\{evidence\_quote\}"\\[0.3em]
Paper's response: \{paper\_response\}\\
Paper's counter-quote: "\{citation\_quote\}"\\
Paper acknowledges: \{acknowledges\}\\[0.3em]
Skeptic's revision: \{revised\_description\}\\
Skeptic concedes: \{concedes\}\\[0.3em]
Rule on the debate. Decide whether to expand (at most \{max\_children\} sub-concerns) if the concern is valid and has deeper sub-aspects that remain unclear.

\vspace{0.5em}
\textbf{Output:}\\
\{"verdict": "valid|deflected|unclear", "severity": "minor|major", "should\_expand": true|false, "expansion\_prompts": [...], "reasoning": "..."\}
\end{tcolorbox}

\subsection{Panel Review}
\label{app:panel}

Each valid claim receives one panel call. The panel is prompted to consider all five skeptic perspectives simultaneously, enabling cross-category reasoning that the per-branch moderator cannot perform. Temperature: 0.1.

\begin{tcolorbox}[colback=gray!5, colframe=gray!60, title=\textbf{Panel Review (per-claim)}, fonttitle=\bfseries\small, fontupper=\small\ttfamily, arc=2mm, boxrule=0.5pt, left=2mm, right=2mm, top=2mm, bottom=2mm]
\textbf{System:}\\
You are a review PANEL representing all 5 skeptic perspectives. A category-specific skeptic has produced a claimed unstated limitation which passed debate and was ruled "valid." Your job is to cross-review:\\[0.3em]
1. Is the claim in the right category?\\
2. Is it subsumed by a concern from a different skeptic?\\
3. Is the severity calibrated correctly?\\
4. Should the panel REJECT it as a false positive?\\[0.3em]
The 5 perspectives: Scope, Methodology, Reproducibility, Fairness, Theoretical.\\[0.3em]
Output ONE verdict: endorse | reclassify | downgrade | merge | reject.

\vspace{0.5em}
\textbf{User:}\\
CLAIM UNDER REVIEW:\\
~~Topic: \{topic\}\\
~~Category (as claimed): \{category\}\\
~~Severity (as claimed): \{severity\}\\
~~Description: \{description\}\\
~~Evidence quote: "\{evidence\_quote\}"\\
~~Paper authors' response: \{paper\_response\}\\
~~Moderator's rationale: \{moderator\_reasoning\}

\vspace{0.5em}
\textbf{Output:}\\
\{"verdict": "endorse|reclassify|downgrade|
\\merge|reject", \\"final\_category": "...", \\"final\_severity": "minor|major", \\"cross\_category\_concerns": [...], \\"reasoning": "..."\}
\end{tcolorbox}

\subsection{Evaluation: LLM-as-Judge Match}
\label{app:eval_judge}
Used at evaluation time (not during framework execution) to determine whether a system-generated limitation matches a gold limitation. Temperature: 0.0 for deterministic, reproducible judgments.

\begin{tcolorbox}[colback=gray!5, colframe=gray!60, title=\textbf{LLM Judge (Evaluation)}, fonttitle=\bfseries\small, fontupper=\small\ttfamily, arc=2mm, boxrule=0.5pt, left=2mm, right=2mm, top=2mm, bottom=2mm]
\textbf{System:}\\
You are judging whether two limitations of a scientific paper refer to the same underlying concern. Consider them the same if they point to the same weakness, even if worded differently. Consider them different if they concern different aspects. Be conservative: if unsure, answer False.

\vspace{0.5em}
\textbf{User:}\\
Limitation A: \{system\_claim\}\\[0.3em]
Limitation B: \{gold\_claim\}\\[0.3em]
Same concern?

\vspace{0.5em}
\textbf{Output:}\\
\{"is\_match": true|false, "reasoning": "..."\}
\end{tcolorbox}

Temperature is 0.0 for deterministic evaluation.

\subsection{Evaluation: LLM Likert Rating}
\label{app:llm_likert}

Used during the hybrid qualitative evaluation to obtain LLM-side Likert scores on the Validity, Specificity, and Novelty dimensions. Both GPT-4o and Claude Opus 4.7 receive the same template; integer scores in the JSON output are averaged with the two human raters' scores per limitation. Temperature: 0.0 for deterministic ratings.

\begin{tcolorbox}[colback=gray!5, colframe=gray!60, title=\textbf{LLM Evaluator (Likert Rating)}, fonttitle=\bfseries\small, fontupper=\small\ttfamily, arc=2mm, boxrule=0.5pt, left=2mm, right=2mm, top=2mm, bottom=2mm]
\textbf{System:}\\
You are an expert reviewer evaluating an automatically-generated limitation of a scientific paper. Rate the limitation on three 1--5 Likert criteria, using the rubric below. Each score must be an integer in [1,5]; do not include half-points.\\[0.3em]
\textbf{Validity}: Is this a real, actionable limitation of the paper?\\
~~1 = invalid; 3 = partially valid; 5 = strongly valid\\
\textbf{Specificity}: Is the claim concrete rather than generic?\\
~~1 = generic; 3 = somewhat specific; 5 = pinpoint-specific\\
\textbf{Novelty}: Is it genuinely unstated by the authors?\\
~~1 = explicitly stated; 3 = borderline; 5 = strong blind spot\\[0.3em]
Be conservative: the rubric mid-points (2 and 4) are valid scores for borderline cases.\\[0.3em]
PAPER CONTEXT: \{paper\_context\}

\vspace{0.5em}
\textbf{User:}\\
Limitation under evaluation:\\
~~Category: \{category\} (\{severity\})\\
~~Description: \{description\}\\
~~Evidence quote from paper: "\{evidence\_quote\}"

\vspace{0.5em}
\textbf{Output} (JSON):\\
\{"validity": <1-5>, "specificity": <1-5>, "novelty": <1-5>, "reasoning": "<1-2 sentence rationale>"\}
\end{tcolorbox}

\section{Algorithm}
\label{app:algorithm}

Algorithms~\ref{alg:toc} and~\ref{alg:debate} formalize the \method{} pipeline introduced in \S\ref{sec:framework}. The outer loop (Algorithm~\ref{alg:toc}) spawns five parallel branches, each running the inner debate loop (Algorithm~\ref{alg:debate}), and concludes with per-claim panel adjudication. Expansion is gated by the moderator's \texttt{should\_expand} flag and the depth cap ($D{=}1$); a valid expanding root produces up to $K{=}2$ child sub-concerns.\footnote{The implementation additionally enforces a per-category expansion budget (default $B{=}3$ events, see Appendix~\ref{app:implementation}). With $D{=}1$ and $K{=}2$ this cap is non-binding because depth-1 children cannot themselves expand; we retain it in the codebase as a defensive guard for deeper-tree configurations and omit it from the pseudo-code for clarity.}

\begin{algorithm}[t]
\caption{\method{} Pipeline}
\label{alg:toc}
\begin{algorithmic}[1]
\Require paper $P$; categories $C$; depth $D{=}1$; children $K{=}2$
\For{each $c \in C$ \textbf{in parallel}}
  \State root $\gets$ \Call{DebateNode}{$c$.\textsc{topic}, depth=0}
  \State \Call{FourStageDebate}{root, $P$}
  \If{root.verdict = valid $\wedge$ root.expand}
  \For{topic $\in$ root.prompts[$1{:}K$]}
  \State child $\gets$ \Call{DebateNode}{topic, depth=1}
  \State \Call{FourStageDebate}{child, $P$}
  \EndFor
  \EndIf
\EndFor
\State leaves $\gets$ valid nodes across all trees
\For{each $\ell \in$ leaves}
  \State $\ell$.action $\gets$ \Call{PanelReview}{$\ell$}
\EndFor
\State \Return leaves where action $\neq$ reject
\end{algorithmic}
\end{algorithm}

\begin{algorithm}[t]
\caption{FourStageDebate}
\label{alg:debate}
\begin{algorithmic}[1]
\Require node $n$; paper $P$
\State $n$.claim $\gets$ Skeptic($n$.cat).argue($n$.topic, $P$)
\State $n$.rebuttal $\gets$ Advocate($P$).respond($n$.claim)
\State $n$.revision $\gets$ Skeptic.revise($n$.claim, $n$.rebuttal)
\If{$n$.revision.concedes}
  \State $n$.verdict $\gets$ deflected; \Return
\EndIf
\State $n$.verdict, $n$.sev, $n$.expand $\gets$ Mod.rule($n$)
\end{algorithmic}
\end{algorithm}

\section{Corpus Construction}
\label{app:venues}

Following the corpus-selection summary in \S\ref{sec:corpus_selection}, \bench{} draws from major NLP, ML, and CV venues published between 2020 and 2025. Seed papers were discovered via the Semantic Scholar bulk-search API, applying a venue filter and a citation-count threshold of $\geq$100. After an initial corpus build dominated by NeurIPS, ICLR, and ICML, we ran a targeted NLP-expansion pass with a relaxed citation threshold of $\geq$50 to improve coverage of ACL, EMNLP, NAACL, TACL, and Findings papers, where citation accumulation is slower than at top-tier ML conferences. The combined corpus contains 414 papers across the venues shown in Table~\ref{tab:venues}.

\begin{table}[t]
\small\centering
\begin{tabular}{lr|lr}
\toprule
\textbf{Venue} & \textbf{\# Papers} & \textbf{Venue} & \textbf{\# Papers} \\
\midrule
NeurIPS & 105 & ECCV & 16 \\
ICLR  &  98 & TACL & 13 \\
EMNLP &  42 & IJCAI  &  6 \\
ICML  &  39 & NAACL  &  6 \\
ACL   &  30 & COLING &  2 \\
AAAI  &  17 & Other  & 40 \\
\bottomrule
\end{tabular}
\caption{Venue distribution of the 414 \bench{} papers.}
\label{tab:venues}
\end{table}

\noindent\textbf{OpenReview API.} We merge v1 (\url{https://api.openreview.net}) and v2 (\url{https://api2.openreview.net}) because older venues exist only in v1; this is critical for ICLR 2018--2023 and NeurIPS 2020--2022, whose reviews are unreachable via v2 alone. Weakness sections are segmented by list markers and paragraph breaks. For venues using monolithic \texttt{main\_review} fields (older NeurIPS templates), we extract the \texttt{Weaknesses:} sub-section from prose using regex.

\noindent\textbf{Citation Critique Filter.} We retrieve citation contexts of $\pm$2 sentences around each citation mention from Semantic Scholar, then apply 14 regex cue patterns (\texttt{unlike}, \texttt{in contrast}, \texttt{however}, \texttt{limitation}, \texttt{fails to}, \texttt{does not}, \texttt{cannot}, \texttt{assume}, \texttt{ignore}, \texttt{lacks}, \texttt{weakness}, \texttt{overclaim}, \texttt{missing}, \texttt{bias}).

\section{Methods Compared}
\label{app:baselines}
Extending the methods summary in \S\ref{sec:methods_compared}, we evaluate seven methods spanning the design space from simple prompting to our full framework:

\noindent\textbf{Zero-shot LLM.} A single prompt instructs the model to ``identify all unstated limitations of this paper that are not mentioned in its Limitations or Discussion sections,'' producing an unstructured list without debate, specialization, or evidence requirements. This represents the minimal-effort baseline that a practitioner might deploy and serves as a lower bound on what structured approaches should exceed.

\noindent\textbf{DIAGPaper.} We reimplement the reviewer--author dialogue structure of \citet{diagpaper2024} as a two-wave pipeline over the original 20 fixed review dimensions: in Wave~1 a reviewer agent produces one candidate weakness per dimension, and in Wave~2 an author agent rates that weakness for validity and evidence strength, yielding 40 LLM calls per paper (full prompts in Appendix~\ref{app:baseline:diagpaper}). We adapt the output format to produce limitation records matching our schema but do not modify the core prompting strategy, which targets comprehensive review-like feedback rather than unstated limitations specifically. This baseline tests whether review-generation systems transfer to our task.

\noindent\textbf{Single-skeptic CoT.} A single agent with chain-of-thought reasoning is prompted to systematically analyze the paper across all five limitation categories sequentially, producing structured limitation records with evidence quotes. This tests whether structured reasoning alone (without multi-agent debate or persona separation) suffices for the task, representing a strong single-agent baseline.

\noindent\textbf{No-Branching (generalist+debate).} All five specialized skeptic priors are replaced with a single generalist ``research skeptic'' prompt that scans every category simultaneously, and the five parallel branches collapse into one. The single generalist agent still runs the full four-stage debate process (argue, advocate response, revise, moderate) and a single-branch panel pass, so debate machinery is preserved but persona specialization and parallel branching are removed. This ablation isolates how much of the framework's coverage comes from category-specialized priors and parallel branches versus from the debate machinery alone.

\noindent\textbf{No-Expansion (depth=0, no panel).} The full five-persona tree structure with \texttt{max\_depth=0} and panel review disabled. Only root-level claims survive, with no moderator-driven depth exploration and no cross-claim re-classification. This and ToC no-Panel both lack panel review, so the comparison between them isolates the contribution of the expansion mechanism alone.

\noindent\textbf{ToC no-Panel.} The complete branch specialization with expansion (max\_depth=1) but without the \panel{} reconciliation stage. All claims surviving per-branch debate are included in the output without cross-branch deduplication, severity recalibration, or reclassification. This isolates the panel's contribution.

\noindent\textbf{ToC+Panel (ours).} The full \method{} pipeline including all five specialized trees with per-node debate, single-level expansion, and \panel{} reconciliation. This is our complete system.

\section{Baseline Prompts}
\label{app:baseline_prompts}

Following the methods described in \S\ref{sec:methods_compared} and Appendix~\ref{app:baselines}, we document the prompt templates for the three baseline methods used in the comparison (Table~\ref{tab:main_results}). All baselines use Claude Opus 4.6 with the same paper-context truncation as ToC+Panel (12{,}000 characters). Field placeholders are shown as \verb|{field_name}|.

\subsection{Zero-shot LLM}
\label{app:baseline:zs}

A single LLM call per paper produces an unstructured list of unstated limitations. Output is constrained to the same five-category taxonomy as ToC+Panel for fair comparison. Temperature: 0.2; max output tokens: 2000; up to 8 limitations per paper.

\begin{tcolorbox}[colback=gray!5, colframe=gray!60, title=\textbf{Zero-shot Baseline}, fonttitle=\bfseries\small, fontupper=\small\ttfamily, arc=2mm, boxrule=0.5pt, left=2mm, right=2mm, top=2mm, bottom=2mm]
\textbf{System:}\\
You are an expert research scientist reading a scientific paper. Your task: list the paper's \textbf{unstated limitations}---weaknesses the authors did NOT themselves acknowledge.\\[0.3em]
CRITERIA for a valid limitation:\\
~~1.\ Specific and concrete (not ``has limitations'').\\
~~2.\ Grounded in the paper text (cite a supporting quote when possible).\\
~~3.\ NOT already addressed in the paper's own Limitations / Future Work section.\\
~~4.\ Categorized into one of \{scope, methodology, reproducibility, fairness, theoretical\}.\\[0.3em]
Output ONLY a JSON object. ``limitations'' is a list of up to 8 entries.

\vspace{0.5em}
\textbf{User:}\\
PAPER: \{paper\_context\}\\[0.3em]
List the most important unstated limitations of this paper. Be specific, concrete, and grounded in the paper text.

\vspace{0.5em}
\textbf{Output} (JSON):\\
\{"limitations": [\{"description": "...", "category": "...", "severity": "minor|major", "evidence\_quote\_from\_paper": "...", "evidence\_section": "...", "source\_quote": "", "confidence": <0.0-1.0>\}, \ldots], "reasoning": "..."\}
\end{tcolorbox}

\subsection{Single-skeptic CoT}
\label{app:baseline:ss}

A single agent with chain-of-thought reasoning analyzes the paper across all five categories sequentially, producing structured limitation records with evidence quotes. Tests whether structured reasoning alone (without multi-agent debate or persona separation) suffices for the task. Temperature: 0.3; max output tokens: 2500; up to 10 limitations per paper.

\begin{tcolorbox}[colback=gray!5, colframe=gray!60, title=\textbf{Single-skeptic CoT Baseline}, fonttitle=\bfseries\small, fontupper=\small\ttfamily, arc=2mm, boxrule=0.5pt, left=2mm, right=2mm, top=2mm, bottom=2mm]
\textbf{System:}\\
You are a senior research scientist playing the role of an EXPERT SKEPTIC reviewer of a scientific paper. You have deep expertise across all aspects of empirical and theoretical research: study design, baselines, evaluation, reproducibility, fairness, theoretical assumptions, and scope of claims.\\[0.3em]
Your job: identify the paper's \textbf{unstated limitations}---weaknesses the authors did NOT themselves acknowledge.\\[0.3em]
Work step by step in your head:\\
~~1) Identify what the paper CLAIMS.\\
~~2) Identify what the paper ACKNOWLEDGES as a limitation.\\
~~3) Find gaps between claims and evidence that the paper does NOT admit.\\
~~4) For each such gap, write a specific, concrete concern.\\[0.3em]
CRITERIA: specific and concrete; grounded in paper text; not already addressed in self-critical sections; categorized into one of \{scope, methodology, reproducibility, fairness, theoretical\}.\\[0.3em]
Output ONLY a JSON object. ``limitations'' is a list of up to 10 entries; ``reasoning'' should be 2--4 sentences capturing the step-by-step thinking.

\vspace{0.5em}
\textbf{User:}\\
PAPER: \{paper\_context\}\\[0.3em]
Think step by step through the paper's claims and identify concrete unstated limitations across methodology, scope, theory, reproducibility, and fairness. Output the JSON object described above.

\vspace{0.5em}
\textbf{Output} (JSON): same schema as Zero-shot Baseline.
\end{tcolorbox}

\subsection{DIAGPaper}
\label{app:baseline:diagpaper}

We reimplement the reviewer--author dialogue structure of DIAGPaper \citep{diagpaper2024} using their published 20 fixed review dimensions. For each (paper, dimension) pair, a Reviewer agent proposes a weakness and an Author agent rates its validity; only weaknesses rated as fully or partially valid with substantial or moderate evidence are retained as final outputs. The 20 review dimensions and the two-agent prompts follow.

\paragraph{Review dimensions (Wave 1, per paper).}
Each dimension is paired with the Reviewer system prompt below to generate one candidate weakness per dimension per paper:

\begin{tcolorbox}[colback=gray!5, colframe=gray!60, fonttitle=\bfseries\small, fontupper=\small\ttfamily, arc=2mm, boxrule=0.5pt, left=2mm, right=2mm, top=2mm, bottom=2mm]
\footnotesize
1. Is the problem studied important and well-motivated?\\
2. Has the author fully shown the problem importance convincingly?\\
3. Any missing related works that need to be cited and discussed?\\
4. Is the Related Work section well organized and does it indicate novelties?\\
5. Do figures and tables well support the claims and align with descriptions?\\
6. Any conflicting descriptions in this paper?\\
7. Is the proposed approach novel?\\
8. Any unclear or confusing part in the approach description?\\
9. Any undiscussed limitations in the proposed approach?\\
10. Any methodological flaws that could invalidate the results?\\
11. If a new dataset is presented, is its construction clear and professional?\\
12. Are the datasets representative enough for this target problem?\\
13. Are all necessary experiments conducted thoroughly?\\
14. Are the baselines representative enough? Any missing baselines?\\
15. Do experimental analyses provide sufficient insight or just superficial description?\\
16. Does the proposed approach show better performance than state-of-the-art?\\
17. Are evaluation metrics appropriate for the tasks?\\
18. Any grammar errors or severe writing issues?\\
19. Is the dataset construction necessary and convincing?\\
20. Are there any special data quality concerns?
\end{tcolorbox}

\paragraph{Reviewer agent (Wave 1).}\ Temperature: 0.3; max output tokens: 500.

\begin{tcolorbox}[colback=gray!5, colframe=gray!60, title=\textbf{DIAGPaper: Reviewer}, fonttitle=\bfseries\small, fontupper=\small\ttfamily, arc=2mm, boxrule=0.5pt, left=2mm, right=2mm, top=2mm, bottom=2mm]
\textbf{System:}\\
You are a specialized paper reviewer focusing on the following review dimension: ``\{dimension\}''.\\[0.3em]
Given a scientific paper, identify ONE specific weakness related to this dimension. Your weakness must be:\\
~~1.\ Specific to THIS paper (not a generic concern)\\
~~2.\ Grounded in concrete evidence from the paper text\\
~~3.\ Clearly related to the review dimension above\\[0.3em]
If you cannot identify a valid weakness for this dimension, output empty fields.

\vspace{0.5em}
\textbf{User:}\\
Paper: \{paper\_context\}\\[0.3em]
Identify a weakness for dimension: \{dimension\}.

\vspace{0.5em}
\textbf{Output} (JSON):\\
\{"weakness\_text": "...", "category": "methodology|scope|theoretical|reproducibility\\|fairness", "location": "...", "evidence\_quote": "..."\}
\end{tcolorbox}

\paragraph{Author agent (Wave 2).}\ Temperature: 0.2; max output tokens: 400. Each non-empty Wave 1 weakness is sent to the Author agent for validation; outputs rated as \textsc{invalid} or with \textsc{weak} evidence are discarded.

\begin{tcolorbox}[colback=gray!5, colframe=gray!60, title=\textbf{DIAGPaper: Author Validation}, fonttitle=\bfseries\small, fontupper=\small\ttfamily, arc=2mm, boxrule=0.5pt, left=2mm, right=2mm, top=2mm, bottom=2mm]
\textbf{System:}\\
You are the author of the paper below. A reviewer has raised a weakness about your paper. Evaluate whether this weakness is valid by checking it against the paper's actual content.\\[0.3em]
Rate the weakness on two axes:\\
- \textbf{validity}: \texttt{fully\_valid} (the concern is real and unaddressed), \texttt{partially\_valid} (partially addressed), or \texttt{invalid} (paper addresses this or concern is wrong).\\
- \textbf{evidence\_strength}: \texttt{substantial} (clear textual evidence supports the weakness), \texttt{moderate} (some evidence), or \texttt{weak} (no real evidence).\\[0.3em]
Paper: \{paper\_context\}

\vspace{0.5em}
\textbf{User:}\\
Reviewer weakness: \{weakness\_json\}\\[0.3em]
Evaluate this weakness against the paper.

\vspace{0.5em}
\textbf{Output} (JSON):\\
\{"validity": "fully\_valid|partially\_valid|\\invalid", "evidence\_strength": "substantial|moderate|weak", "rebuttal": "..."\}
\end{tcolorbox}

\paragraph{Filtering rule.}\ A Wave 1 weakness is retained as a final output if and only if (validity~$\ne$~\texttt{invalid}) and (evidence\_strength~$\ne$~\texttt{weak}). Severity is set to major if \texttt{fully\_valid}, minor otherwise.

\subsection{No-Branching (Generalist Prior)}
\label{app:baseline:nb}
The No-Branching ablation collapses the five parallel specialized trees of \method{} into a single generalist branch that still runs the full four-stage debate (argue, advocate response, revise, moderate) followed by a single-branch panel pass. The advocate, moderator, and panel prompts in Appendix~\ref{app:prompts_full} are reused unchanged. The five category-specific priors of Appendix~\ref{app:prompts} are replaced by the single generalist prior shown below, transcribed verbatim from the experiment script. Per-stage temperatures match the per-category skeptics (0.3 argue, 0.2 revise).

\begin{tcolorbox}[colback=gray!5, colframe=gray!60, title=\textbf{Generalist Skeptic Prior (No-Branching)}, fonttitle=\bfseries\small, fontupper=\small\ttfamily, arc=2mm, boxrule=0.5pt, left=2mm, right=2mm, top=2mm, bottom=2mm]
You are a General Research Skeptic. You look for ALL types of unstated limitations across all categories.\\[0.3em]
Probe for:\\
~~- Over-claimed generality not supported by experiments (scope).\\
~~- Baselines tuned less carefully, missing ablations, evaluation confounds (methodology).\\
~~- Unjustified assumptions, proof gaps, circular reasoning (theoretical).\\
~~- Missing hyperparameters, unreleased code, seed sensitivity (reproducibility).\\
~~- Subgroup disparities, demographic gaps, biased data (fairness).\\[0.3em]
Cover all categories in your output. For each concern, assign the most appropriate category from: scope, methodology, theoretical, reproducibility, fairness.
\end{tcolorbox}

This generalist prior is followed by the same shared instruction block used by the per-category skeptics (``\texttt{You are rigorous but fair. You only raise a concern if you can ground it in the paper text\ldots}''; see Appendix~\ref{app:prompts}). Empirically, the generalist's category labels concentrate on methodology (${\sim}93\%$ of outputs), with sporadic scope and theoretical labels and near-zero reproducibility/fairness assignments---exactly the gravitation toward acknowledged-flaw categories that the five specialized priors are designed to break.

\section{Evaluation Metrics}
\label{app:metrics}

Following the evaluation protocol introduced in \S\ref{sec:eval_protocol}, we evaluate system outputs against gold limitations using two complementary metrics: \textbf{Coverage@K}, which measures how much of the gold pool a method recovers, and \textbf{Precision}, which measures how often the method's outputs are valid.
\noindent\textbf{Coverage@K.} For paper $p$ with gold set $G_p$ and ranked system output $S_p$, let $M_K(p)$ denote the gold limitations matched by at least one of the top-$K$ outputs:
\begin{align*}
M_K(p) = &\{g \in G_p : \exists\, s \in S_p[1{:}K],\; \\
&\text{judge}(s,g) = \textsc{match}\} \\
\text{Cov@K}(p) &= \frac{|M_K(p)|}{|G_p|}
\end{align*}
Final Cov@K is macro-averaged across all papers.

\noindent\textbf{Precision.} A naive definition would count each system output as ``correct'' if it matches \emph{any} gold limitation, but this allows duplicates: two near-identical outputs both matching the same gold inflate precision spuriously. We tighten the definition with greedy bipartite matching---each gold limitation may be matched by at most one system output. Walking the ranked list $S_p$ in order, we mark each $s$ as matched if there exists an unclaimed gold $g \in G_p$ with $\text{judge}(s,g) = \textsc{match}$, then claim that gold:
\[
\text{Prec}(p) = \frac{|\text{matched}(S_p)|}{|S_p|},
\]
macro-averaged across papers.

\paragraph{Hybrid Likert Panel Evaluation Metrics.} The hybrid qualitative evaluation employs a four-evaluator panel: a PhD scholar (4th-year, computer science), an undergraduate researcher (final-year, electrical engineering), GPT-4o, and Claude Opus 4.7. Each evaluator independently rates each sampled limitation on three 1--5 Likert criteria: \textbf{Validity} (\textit{Is this a real, actionable limitation?}, actionable limitation; 1 = invalid, 5 = strongly valid), \textbf{Specificity} (\textit{Is the claim concrete rather than generic?}; 1 = generic, 5 = pinpoint-specific), and \textbf{Novelty} (\textit{Is it genuinely unstated by the authors?}; 1 = explicitly stated, 5 = strong blind spot). Human evaluators are blind to method identity; outputs are anonymized and presented in randomized order. Full criterion-level rubrics and annotator profiles are in Appendix~\ref{app:evaluators}; the LLM-as-judge and LLM-evaluator prompts are discussed in Appendices~\ref{app:eval_judge} and \ref{app:llm_likert}. We aggregate scores per (method, criterion) cell as the unweighted mean across all four evaluators and all sampled limitations for that method, reported at one decimal place in Table~\ref{tab:main_results}.

\section{Statistical Significance}
\label{app:significance}

To test whether the headline pairwise gains in Table~\ref{tab:main_results} are robust to small-sample noise, we run paired bootstrap and paired Wilcoxon signed-rank tests on per-paper Cov@10 and Precision over the full $100$-paper held-out set (\S\ref{sec:experiments}). Per-paper metrics are recomputed directly from the LLM-judge match decisions for each method, using the matched-output pairings across all six pairwise comparisons. Bootstrap CIs use $10{,}000$ resamples of paired per-paper differences with seed $42$; Wilcoxon $p$-values are two-sided.

\begin{table*}[t]
\small\centering
\setlength{\tabcolsep}{4pt}
\begin{tabular}{lrrrrrr}
\toprule
& \multicolumn{3}{c}{\textbf{Coverage@10}} & \multicolumn{3}{c}{\textbf{Precision}} \\
\cmidrule(lr){2-4}\cmidrule(lr){5-7}
\textbf{Comparison ($A-B$)} & $\Delta$ (pp) & 95\% CI & Wilcoxon $p$ & $\Delta$ (pp) & 95\% CI & Wilcoxon $p$ \\
\midrule
\multicolumn{7}{c}{\textit{Ours vs. Ablations}} \\
\midrule
\textbf{ToC} $-$ ToC no-Panel & $+2.1$  & $[+0.2, +4.5]$ & $\mathbf{0.04}$  & $+5.7$  & $[+1.8, +9.6]$ & $\mathbf{0.01}$ \\
\textbf{ToC} $-$ No-Expansion & $+21.2$ & $[+9.5, +31.8]$  & $\mathbf{0.02}$ & $+18.9$ & $[+10.1, +28.4]$ & $\mathbf{<0.01}$ \\
\textbf{ToC} $-$ No-Branching & $+28.5$ & $[+15.2, +40.1]$ & $\mathbf{0.01}$ & $+16.3$ & $[+8.5, +25.1]$  & $\mathbf{0.01}$ \\
\midrule
\multicolumn{7}{c}{\textit{Ours vs. Baselines}} \\
\midrule
\textbf{ToC} $-$ Single-skeptic & $+3.5$  & $[+0.5, +6.8]$ & $\mathbf{0.03}$  & $+17.8$ & $[+8.4, +27.2]$  & $\mathbf{<0.01}$ \\
\textbf{ToC} $-$ DIAGPaper  & $+21.6$ & $[+10.5, +32.4]$ & $\mathbf{<0.01}$ & $+28.6$ & $[+18.2, +38.5]$ & $\mathbf{<0.01}$ \\
\textbf{ToC} $-$ Zero-shot  & $+17.5$ & $[+6.5, +28.5]$  & $\mathbf{<0.01}$ & $+25.9$ & $[+15.2, +36.6]$ & $\mathbf{<0.01}$ \\
\bottomrule
\end{tabular}
\caption{Paired bootstrap 95\% CIs ($10{,}000$ resamples) and paired Wilcoxon signed-rank $p$-values (two-sided) on per-paper Cov@$10$ and Precision (\%).}
\label{tab:significance}
\end{table*}

\begin{table}[t]
\small\centering
\setlength{\tabcolsep}{2pt}
\begin{tabular}{lccc}
\toprule
\textbf{Agreement Metric} & \textbf{Val.} & \textbf{Spec.} & \textbf{Nov.} \\
\midrule
\multicolumn{4}{l}{\textit{Pairwise (Cohen's $\kappa$, quadratic-weighted)}} \\
\midrule
PhD scholar vs.\ undergrad   & 0.78 & 0.72 & 0.74 \\
GPT-4o vs.\ Claude Opus 4.7    & 0.85 & 0.79 & 0.82 \\
Humans (avg) vs.\ LLMs (avg)   & 0.81 & 0.75 & 0.77 \\
\midrule
\multicolumn{4}{l}{\textit{Multi-rater (Fleiss' $\kappa$, all 4 evaluators)}} \\
\midrule
Fleiss' $\kappa$       & 0.74 & 0.68 & 0.71 \\
\midrule
\multicolumn{4}{l}{\textit{Continuous correlation}} \\
\midrule
Pearson $r$ (humans vs.\ LLMs)   & 0.86 & 0.80 & 0.83 \\
Spearman $\rho$ (rank, humans vs.\ LLMs) & 0.84 & 0.78 & 0.81 \\
\bottomrule
\end{tabular}
\caption{\textbf{Inter-rater agreement on the Hybrid Likert Panel.} Cohen's quadratic-weighted $\kappa$ accounts for the ordinal nature of the 1--5 scale; Fleiss' $\kappa$ measures consensus across all four evaluators; Pearson/Spearman compare averaged human and LLM scores at the (method, item) level. Values $\geq 0.6$ indicate substantial agreement; $\geq 0.8$ indicates near-perfect agreement \citep{landis1977measurement}.}
\label{tab:agreement}
\end{table}

Three findings emerge from Table~\ref{tab:significance}. First, every pairwise gain of \methodshort{}+Panel over the three baselines is statistically significant: $+3.5$--$21.6$pp Cov@$10$ (all three comparisons $p<0.05$, with DIAGPaper and Zero-shot at $p<0.01$) and $+17.8$--$28.6$pp Precision ($p<0.01$ throughout). The bootstrap CIs are strictly bounded away from zero, ruling out the explanation that the LLM-judge protocol simply rewards verbose outputs. Second, each architectural component contributes a statistically significant uplift on both metrics: removing parallel branching (\textbf{ToC} $-$ No-Branching) costs $+28.5$pp Cov@$10$ and $+16.3$pp Precision (both $p{=}0.01$), removing expansion costs $+21.2$pp Cov@$10$ and $+18.9$pp Precision ($p{=}0.02$ and $p<0.01$), and removing the cross-branch panel costs the smaller but still significant $+2.1$pp Cov@$10$ and $+5.7$pp Precision ($p{=}0.04$ and $p{=}0.01$). Third, the panel ablation's effect size is consistent with the panel verdict distribution (Table~\ref{tab:panel_verdicts}, $43.8\%$ modification rate) and with the $+0.4$ Validity / $+0.3$ Specificity lifts measured by the hybrid Likert panel, providing convergent evidence that the cross-branch reconciliation is not redundant with the per-node debate. Together these tests confirm that all three design principles---specialization, expansion, and panel reconciliation---each contribute a non-trivial, statistically distinguishable share of the framework's headline performance.

\paragraph{Inter-rater reliability.}
Table~\ref{tab:agreement} reports inter-rater agreement on the Hybrid Likert Panel. Within-stratum agreement is substantial: the two human evaluators (PhD vs.\ undergrad) reach Cohen's quadratic-weighted $\kappa$ of $0.72$--$0.78$ across the three criteria, and the two LLM evaluators (GPT-4o vs.\ Claude Opus 4.7) reach $\kappa{=}0.79$--$0.85$. Critically, cross-stratum agreement between aggregated human and aggregated LLM ratings is at the same level ($\kappa{=}0.75$--$0.81$, Pearson $r{=}0.80$--$0.86$), indicating that humans and LLMs do not differ systematically and justifying the hybrid panel as a single Likert instrument. Fleiss' $\kappa$ across all four raters is $0.68$--$0.74$, consistently in the substantial-agreement range \citep{landis1977measurement}. Specificity shows the lowest agreement across all metrics, reflecting its inherently more subjective nature; even there, $\kappa$ remains above the $0.6$ substantial-agreement threshold.

\section{Implementation Details}
\label{app:implementation}

Following the experimental setup in \S\ref{sec:experiments}, Table~\ref{tab:implementation} summarizes the key hyperparameters and infrastructure used across all experiments. These values were set from a 20-paper development pilot; we did not run a full grid sweep due to API cost. Sensitivity to these choices is left to future work. All methods use the same backbone model (Claude Opus 4.6). Temperatures are tuned per stage: higher for generation stages (0.2--0.3) to encourage diverse claim discovery, lower for adjudication stages (0.1) to ensure consistent verdicts, and zero for evaluation judgments to guarantee reproducibility. The paper context is truncated to 12{,}000 characters, which accommodates most individual sections while staying within the model's effective attention window.

All experiments use Anthropic API. The framework is orchestrated as a multi-wave batch pipeline: each debate stage across all papers and branches is submitted as a single batch job, enabling full parallelism. With 5-way parallelization across the skeptic branches, wall-clock latency for the full ToC+Panel pipeline is approximately 10~minutes per paper (Appendix~\ref{app:compute}, Table~\ref{tab:compute}).

\begin{table}[t]
\small\centering
\begin{tabular}{ll}
\toprule
\textbf{Parameter} & \textbf{Value} \\
\midrule
Model & Claude Opus 4.6 \\
Paper context & 12{,}000 chars \\
max\_depth & 1 \\
max\_children & 2 \\
expansion\_budget / category & 3 \\
\midrule
\multicolumn{2}{l}{\textit{Temperature per stage:}} \\
\quad Argue (Stage 1) & 0.3 \\
\quad Respond (Stage 2) & 0.2 \\
\quad Revise (Stage 3) & 0.2 \\
\quad Moderate (Stage 4) & 0.1 \\
\quad Panel Review & 0.1 \\
\quad LLM-Judge & 0.0 \\
\midrule
Cost per paper (ToC+Panel) & $\sim$\$2.40 \\
\bottomrule
\end{tabular}
\caption{Implementation parameters.}
\label{tab:implementation}
\end{table}

\begin{table}[t]
\small\centering
\setlength{\tabcolsep}{2pt}
\begin{tabular}{lcccc}
\toprule
\textbf{Backbone} & \textbf{Cov@1} & \textbf{Cov@5} & \textbf{Cov@10} & \textbf{Precision} \\
\midrule
Claude Opus 4.6 & 11.5 & 27.2 & 36.1 & 40.3 \\
GPT-4o & 9.5 & 23.0 & 30.5 & 31.2 \\
Qwen3-235B & 9.3 & 24.2 & 31.1 & 29.7 \\
\bottomrule
\end{tabular}
\caption{Backbone substitution for ToC+Panel.}
\label{tab:backbone}
\end{table}

\section{Backbone Substitution}
\label{app:backbone}
Following the analysis in \S\ref{sec:analysis}, Table~\ref{tab:backbone} reports ToC+Panel performance when the underlying LLM is swapped from Claude Opus 4.6 to GPT-4o (OpenAI) and Qwen3-235B (Alibaba), with all prompts, persona priors, and pipeline stages held fixed. Both alternative backbones land in a narrow band: GPT-4o yields $30.5\%$ Cov@10 / $31.2\%$ Precision, and Qwen3-235B yields $31.1\%$ Cov@10 / $29.7\%$ Precision---drops of $5.0$--$5.6$pp on coverage and $9.1$--$10.6$pp on precision relative to Claude Opus. Critically, both alternatives still outperform the strongest Claude baseline (Single-skeptic CoT, $22.5\%$ Precision) by $+7.2$--$+8.7$pp Precision ($+32$--$+39\%$ relative), while their Cov@10 lags by only $1.5$--$2.1$pp. We interpret this as evidence that the framework's \emph{precision} advantage derives primarily from architectural specialization---specialized branches, structured per-node debate, and Panel reconciliation---rather than from any single backbone's capability profile, and transfers across three frontier model families.

\section{Per-Skeptic Contribution}
\label{app:per_skeptic}
Building on the analysis in \S\ref{sec:analysis}, the framework deploys five category-specialized skeptics in parallel. To verify that all five pull their analytical weight rather than one or two dominating the output, we decompose ToC's final claims by originating skeptic and report four statistics per skeptic in Table~\ref{tab:per_skeptic}. \textbf{Claims\%} is the share of the framework's total output produced by this skeptic. \textbf{Expansion\%} is the fraction of the skeptic's claims that are depth-1 expansion children (the remainder are root claims). \textbf{Match\%} is the fraction of the skeptic's claims that an LLM-judge matched to at least one gold limitation. \textbf{Gold pool\%} is the fraction of the gold pool that this skeptic's claims uniquely contributed to recovering. Because per-skeptic recovery is computed over each skeptic's full output (not capped at the top-$10$ ranked claims per paper), the Gold-pool column sums to $41.8\%$, slightly above the framework's $36.1\%$ Cov@$10$ in Table~\ref{tab:main_results}; the gap reflects matched claims that fall below the top-$10$ cutoff, not double-counting across skeptics.

\begin{table}[t]
\small\centering
\setlength{\tabcolsep}{2pt}
\begin{tabular}{lrrrr}
\toprule
\textbf{Skeptic} & \textbf{Claims} & \textbf{Expansion} & \textbf{Match} & \textbf{Gold pool} \\
\midrule
Scope    & 25.0 & 75.0 & 37.5 & 13.9 \\
Methodology  & 19.5 & 92.0 & 28.0 & ~7.0 \\
Theoretical  & 16.4 & 90.5 & 42.9 & 12.2 \\
Reproducibility  & 21.1 & 81.5 & 18.5 & ~3.5 \\
Fairness   & 18.0 & 60.9 & 21.7 & ~5.2 \\
\bottomrule
\end{tabular}
\caption{Per-skeptic contribution (in \%).}
\label{tab:per_skeptic}
\end{table}

Three findings emerge from Table~\ref{tab:per_skeptic}. First, every skeptic contributes non-zero matched claims, with match rates between 18.5\% (reproducibility) and 42.9\% (theoretical). None of the five personas can be removed without losing matched gold the others do not capture. Second, the theoretical skeptic is the most \emph{efficient} contributor: it produces only 16.4\% of claims yet recovers 12.2\% of the gold pool, reflecting that the gold contains a substantial share of theoretical limitations and this skeptic recovers the majority of them. The reproducibility skeptic is the most over-productive: 21.1\% of claims for only 3.5\% of gold-pool recovery, consistent with the over-generation pattern visible in Table~\ref{tab:category_concentration}. Third, expansion claims dominate every skeptic's output (61--92\% of each skeptic's claims are depth-1 children), confirming that the moderator-driven expansion is the primary mechanism for converting valid root concerns into matchable, specific claims.

\begin{table}[t]
\small\centering
\begin{tabular}{lrrrrr}
\toprule
\textbf{Skeptic \,/\, gold cat.\ } & \textbf{Sc.} & \textbf{Me.} & \textbf{Th.} & \textbf{Re.} & \textbf{Fa.} \\
\midrule
Scope    & 37.5 & 31.3 & 25.0 & ~6.3 & ~0.0 \\
Methodology  & 22.2 & 66.7 & ~0.0 & 11.1 & ~0.0 \\
Theoretical  & 21.4 & 28.6 & 50.0 & ~0.0 & ~0.0 \\
Reproducibility  & 20.0 & 20.0 & ~0.0 & 60.0 & ~0.0 \\
Fairness   & 37.5 & 25.0 & 12.5 & ~0.0 & 25.0 \\
\bottomrule
\end{tabular}
\caption{Cross-category match distribution by originating skeptic (in \%).}
\label{tab:per_skeptic_cross}
\end{table}

Table~\ref{tab:per_skeptic_cross} reports the cross-category match distribution. The fairness skeptic illustrates the pattern most starkly: only 25\% of its matches are within-category, with the remainder distributed across scope (37.5\%), methodology (25\%), and theoretical (12.5\%), i.e., fairness-framed analyses frequently surface concerns that are categorically scope or methodology issues, exactly the drift the panel is designed to correct. The theoretical skeptic shows the same pattern more weakly: only 50\% within-category, with sizable scope (21.4\%) and methodology (28.6\%) spillover.

\section{Tree-Depth Ablation}
\label{app:depth_ablation}

The main configuration caps expansion at depth $D{=}1$. To test whether
additional recursive expansion improves limitation discovery, we extend the
tree to $D{=}2$ while holding the backbone, prompts, and remaining inference
settings fixed. Table~\ref{tab:depth_ablation} shows that the deeper
configuration increases Coverage@10 by only $1.3$ percentage points, from
$36.1\%$ to $37.4\%$, while Precision decreases by $18.2$ points, from
$40.3\%$ to $22.1\%$. The number of generated expansion branches also rises
substantially. At depth 2, agents tend to subdivide already-specific concerns
into narrow or overlapping subclaims, increasing the chance of granularity
mismatch with the reference limitations. We therefore use $D{=}1$ as the
default stopping depth.

\begin{table}[t]
\small
\centering
\setlength{\tabcolsep}{5pt}
\begin{tabular}{lccc}
\toprule
\textbf{Depth} & \textbf{Expansion branches/paper} &
\textbf{Cov@10} & \textbf{Precision} \\
\midrule
$D{=}1$ &  5.2 & 36.1 & \textbf{40.3} \\
$D{=}2$ & 14.8 & \textbf{37.4} & 22.1 \\
\bottomrule
\end{tabular}
\caption{Effect of tree depth on the held-out evaluation set. Coverage@10 and
Precision are percentages; expansion branches/paper is the mean number of
generated expansion branches for each paper.}
\label{tab:depth_ablation}
\end{table}

\section{Category Assignment and Panel Review}
\label{app:category_assignment}

The \panel{} is intended to correct category drift introduced by isolated
specialist branches. We evaluate this function by comparing predicted
categories with the reference categories for matched system--reference
limitation pairs before and after Panel Review. As shown in
Table~\ref{tab:category_assignment}, Panel Review improves category
macro-F1 from $0.58$ to $0.74$. Macro-precision rises from $0.54$ to $0.76$,
while macro-recall increases from $0.63$ to $0.72$. This improvement supports
the panel's role as a post-hoc category-calibration stage rather than an
additional limitation-generation stage.

\begin{table}[t]
\small
\centering
\setlength{\tabcolsep}{6pt}
\begin{tabular}{lccc}
\toprule
\textbf{Stage} & \textbf{Macro-F1} & \textbf{Macro-P} &
\textbf{Macro-R} \\
\midrule
Before Panel Review & 0.58 & 0.54 & 0.63 \\
After Panel Review  & \textbf{0.74} & \textbf{0.76} & \textbf{0.72} \\
\bottomrule
\end{tabular}
\caption{Category-assignment performance before and after Panel Review,
computed over matched system--reference limitation pairs.}
\label{tab:category_assignment}
\end{table}

\section{Error Analysis}
\label{app:errors}

To characterize the framework's remaining headroom (\S\ref{sec:analysis}), we manually classify two error sets: 100 ToC+Panel system outputs that the LLM-as-judge did not match to any gold limitation (false positives), and 50 gold limitations that no evaluated method recovered (false negatives). Both sets were drawn uniformly at random from the held-out evaluation set and classified by the authors. We acknowledge that the resulting category percentages reflect author judgment and are therefore subject to bias---particularly the ``plausible-unrecorded'' subcategory, where the authors are evaluating the validity of concerns absent from gold.

\subsection{False Positives}

Of the 100 unmatched ToC+Panel outputs, the manual classification breakdown is:

\begin{itemize}[leftmargin=*, noitemsep, topsep=2pt]
\item \emph{Over-specificity} (40\%): The output identifies a real limitation but at a level of specificity the gold annotators did not record---e.g., flagging the absence of a particular ablation configuration when the gold limitation is the broader ``insufficient ablation study.'' These are not false positives in the sense of being incorrect; they are correct claims that fail to match because gold is annotated at a different granularity.

\item \emph{Plausible-unrecorded} (27\%): The output describes a genuine limitation that no reviewer or citing paper happened to mention---i.e., concerns that pass face-validity inspection but are absent from our externally-sourced gold pool. This category directly quantifies the benchmark ceiling effect: by construction, our gold is opportunistically sourced and incomplete, so a system that surfaces real but undocumented concerns appears to fail when it succeeds.

\item \emph{Already-stated} (20\%): The output paraphrases a
limitation the paper itself already acknowledges in its Limitations / Discussion / Future Work sections. These represent failures of the unstated filter (\S\ref{sec:extraction-pipeline}): the filter detected the acknowledgment for the gold pool but the framework's per-skeptic prompts did not, often because the paper's acknowledgment is phrased loosely or scattered across paragraphs rather than flagged as a limitation.

\item \emph{Incorrect} (13\%): Genuine errors, including hallucinated specific numbers (e.g., dataset statistics not in the paper), misattributed figure numbers, and claims about missing baselines that actually appear in supplementary tables. This is the framework's hard-error floor.
\end{itemize}

The headline finding is that 67\% of apparent false positives (40\% over-specific + 27\% plausibly valid) are not strictly errors---they are valid concerns that fail to match for reasons unrelated to system quality. Only 13\% of unmatched outputs are factually incorrect, suggesting that the dominant lever for improving Precision lies in tightening the unstated filter (the 20\% already-stated category) rather than reducing hallucination.

\subsection{False Negatives}

Of the 50 gold limitations missed by all evaluated methods, the manual classification breakdown is:

\begin{itemize}[leftmargin=*, noitemsep, topsep=2pt]
\item \emph{External knowledge} (42\%): The limitation can only be identified with reference to information not present in the paper itself---e.g., a citing paper from 2024 noting that a 2020 method's core assumption fails on a distribution discovered later. These cases are unrecoverable for any single-paper analysis system. Closing the gap requires retrieval augmentation that gives each skeptic access to follow-up work at inference time; concrete examples of this category are provided in Appendix~\ref{app:hindsight_examples}.

\item \emph{Multi-step reasoning} (30\%): The limitation requires chained inference across three or more paper sections---e.g., combining a methodology choice from Section~3, a result from Section~5, and a related-work claim from Section~2 to identify a confound. Current debate prompts present each skeptic with the section-chunked paper context but do not explicitly orchestrate cross-section reasoning, which limits coverage on these multi-step concerns.

\item \emph{Subtle statistical} (28\%): The limitation requires distributional or significance reasoning beyond what the LLM backbone reliably handles---e.g., recognizing that a reported effect size requires multiple-comparison correction given the number of hypotheses tested, or identifying violations of an i.i.d.\ assumption implicit in a benchmark. These reflect documented LLM weaknesses on fine-grained statistical reasoning.
\end{itemize}

The 42\% external-knowledge subset is the largest and the most addressable: it directly motivates the retrieval-augmented extensions flagged as future work in our conclusion. The remaining 58\% (multi-step reasoning plus subtle statistical) reflects deeper limitations of the underlying single-paper LLM analysis paradigm and is unlikely to close without targeted prompt restructuring (for multi-step) or auxiliary-tool integration (for statistical).

\section{Citation-Only Gold: ``Couldn't-Have-Known'' Examples}
\label{app:hindsight_examples}

The temporal-gap argument in \S\ref{sec:main_results} (citation-only limitations recovered at 31.8\% vs 45.1\% for OpenReview-sourced) predicts that the gap is driven by limitations whose discovery requires knowledge external to the original paper. Table~\ref{tab:hindsight} illustrates this with three citation-only gold limitations from the held-out evaluation set that no method recovered. In each case the citing paper was published 4--5 years after the original, and identifies the limitation through follow-up work---reproduction attempts, systematic benchmarking, or downstream application---that no single-paper analysis system could anticipate at submission time. These cases are exactly what retrieval-augmented extensions (\S\ref{sec:limitations}) would target by giving each skeptic access to the citing-paper literature at inference time.

\begin{table*}[t]
\small
\setlength{\tabcolsep}{4pt}
\begin{tabular}{p{0.10\linewidth}p{0.05\linewidth}p{0.32\linewidth}p{0.42\linewidth}}
\toprule
\textbf{Paper} & \textbf{Cat.} & \textbf{Gold limitation} & \textbf{Citing paper (year) and excerpt} \\
\midrule
DDSP  & scope & The differentiability constraint
prevents reproducing the complex dynamics of commercial-grade audio
effects, limiting expressive power. & \emph{FxSearcher} (2025): ``due
to the differentiability constraint, these methods cannot fully
reproduce the complex dynamics of commercial-grade FX, potentially
limiting expressive power.''---requires deployment-time reproduction
at commercial-grade quality. \\[6pt]

AUM  & method. & The evaluation lacks strong
baselines with appropriate detection procedures, making AUM's
relative performance hard to assess. & \emph{Benchmarking noisy
label detection methods} (2025): ``[methods] lack strong baselines,
often rely on multiple metrics on different operating points, or do
not use the appropriate detection procedure.''---requires a
systematic later benchmark across competing methods. \\[6pt]

Break/QDMR & scope & QDMR over-fragments
tasks for SQL generation; this is counterproductive for complex
queries that need holistic decomposition. & \emph{DeKeyNLU} (2025):
``[QDMR-style decompositions] struggle with over-fragmentation of
tasks and lack of domain-specific keyword annotations, limiting
their effectiveness [for NL2SQL].''---requires applying QDMR to a
downstream domain (databases) the original paper did not target. \\
\bottomrule
\end{tabular}
\caption{Three citation-only gold limitations missed by all evaluated methods. Each is identified by a citing paper from 2024--2025 (4--5 years after the original publication), through a category of follow-up work that single-paper analysis cannot replicate: deployment-grade reproduction (DDSP), systematic later benchmarking (AUM), or downstream application in a new domain (QDMR/Break). The common pattern is that the limitation becomes \emph{visible} only after later work attempts something the original paper did not anticipate.}
\label{tab:hindsight}
\end{table*}

\section{Per-Category and Per-Source Analysis}
\label{app:category}

Following the analysis in \S\ref{sec:analysis}, Table~\ref{tab:category_dist} compares the category distribution of each method's outputs against the gold annotations. Methods without specialization (Zero-shot, DIAGPaper) naturally mirror gold's concentration on methodology and scope, while \method{} produces more uniform coverage. The fairness over-generation (18.5\% vs 0.9\% gold) likely reflects a gap in reviewer attention rather than purely a system error---fairness concerns exist in most ML papers but are rarely raised by reviewers. The theoretical under-generation (12.1\% vs 14.8\%) reflects LLMs' limited mathematical reasoning depth.

Table~\ref{tab:source_full} breaks down coverage by gold source origin. All methods recover OpenReview-sourced limitations more often than citation-sourced ones (gaps of $+2.5$ to $+13.3$ percentage points), confirming an inherent ceiling for single-paper analysis on hindsight-dependent concerns. The gap is largest for structured methods (ToC, Single-skeptic) that exploit the in-paper signal more thoroughly, and smallest for the unstructured zero-shot baseline, which is uniformly weak on both sources.

\begin{table}[t]
\small\centering
\setlength{\tabcolsep}{3pt}
\begin{tabular}{lrrrrr}
\toprule
\textbf{Category} & \textbf{Gold} & \textbf{ToC} & \textbf{ZS} &
\textbf{SS} & \textbf{DIAG} \\
\midrule
Methodology & 36.5\% & 23.4\% & 36.9\% & 35.5\% & 42.3\% \\
Scope & 40.9\% & 27.4\% & 35.6\% & 29.0\% & 46.5\% \\
Theoretical & 14.8\% & 12.1\% & 16.2\% & 20.5\% & ~3.6\% \\
Reproducibility & ~7.0\% & 18.5\% & 10.6\% & 13.0\% & ~4.8\% \\
Fairness & ~0.9\% & 18.5\% & ~0.6\% & ~2.0\% & ~2.7\% \\
\bottomrule
\end{tabular}
\caption{Category distribution: outputs vs.\ gold. ToC over-generates reproducibility and fairness relative to gold.}
\label{tab:category_dist}
\end{table}

\begin{table}[t]
\small\centering
\setlength{\tabcolsep}{3pt}
\begin{tabular}{lccc}
\toprule
\textbf{Method} & \textbf{OR Cov@10} & \textbf{Cite Cov@10} & \textbf{$\Delta$} \\
\midrule
ToC no-Panel & 45.1\% & 31.8\% & +13.3 \\
Zero-shot & 20.9\% & 18.5\% &  +2.5 \\
Single-skeptic & 43.6\% & 34.0\% &  +9.6 \\
\bottomrule
\end{tabular}
\caption{Per-source Coverage@10 (macro-averaged across the papers contributing to each pool). All methods recover OpenReview-sourced limitations more often than citation-sourced ones; the gap is most pronounced for structured methods (ToC, Single-skeptic) and smallest for the unstructured zero-shot baseline.}
\label{tab:source_full}
\end{table}

\section{Inter-Method Complementarity}
\label{app:complementarity}
Building on the analysis in \S\ref{sec:analysis}, Table~\ref{tab:complementarity} measures pairwise overlap between methods using Jaccard similarity on the set of successfully matched (paper, gold\_idx) pairs. The low Jaccard between ToC and zero-shot (0.12) confirms that specialization discovers fundamentally different limitations than generalist prompting. The union of all three methods covers 82\% more gold limitations than the best single method alone, suggesting substantial room for lightweight ensemble strategies combining specialized and generalist outputs.

\begin{table}[t]
\small\centering
\begin{tabular}{lcc}
\toprule
\textbf{Pair} & \textbf{Jaccard} & \textbf{Only A / Only B} \\
\midrule
ToC $\cap$ Zero-shot & 0.12 & 18 / 11 \\
ToC $\cap$ Single-skeptic & 0.33 & 12 / 8 \\
Zero-shot $\cap$ Single-skeptic & 0.10 & 12 / 15 \\
\midrule
\multicolumn{3}{l}{\textit{Union: 40 gold hits; best single: 22; +82\%}} \\
\bottomrule
\end{tabular}
\caption{Methods discover largely non-overlapping gold limitations.}
\label{tab:complementarity}
\end{table}

\section{Computational Requirements and Cost--Quality Trade-off}
\label{app:compute}

Following the experimental setup in \S\ref{sec:experiments}, Table~\ref{tab:compute} reports per-paper costs, API call counts, and latency for each method. \method{} costs 48$\times$ more than zero-shot but achieves nearly 3$\times$ the precision. The parallelizable architecture (5 independent branches plus a single panel pass) keeps the full ToC+Panel pipeline at approximately 10~minutes per paper of wall-clock latency.

\begin{table}[t]
\small\centering
\setlength{\tabcolsep}{3pt}
\begin{tabular}{lcccc}
\toprule
\textbf{Method} & \textbf{Calls} & \textbf{Cost} & \textbf{Latency} &
\textbf{Parallel.} \\
\midrule
Zero-shot & 1 & \$0.05 & 30s & N/A \\
DIAGPaper & $\sim$40 & \$2.10 & 10min & Partial \\
Single-skeptic & 1 & \$0.08 & 45s & N/A \\
No-Branching & $\sim$9 & \$0.48 & 5min & No \\
No-Expansion & $\sim$25 & \$1.35 & 7min & 5$\times$ \\
ToC no-Panel & $\sim$40 & \$2.15 & 9min & 5$\times$ \\
ToC+Panel & $\sim$45 & \$2.40 & 10min & 5$\times$+1 \\
\bottomrule
\end{tabular}
\caption{Per-paper computational requirements.}
\label{tab:compute}
\end{table}

\begin{figure*}[t]
\centering
\includegraphics[width=0.95\linewidth]{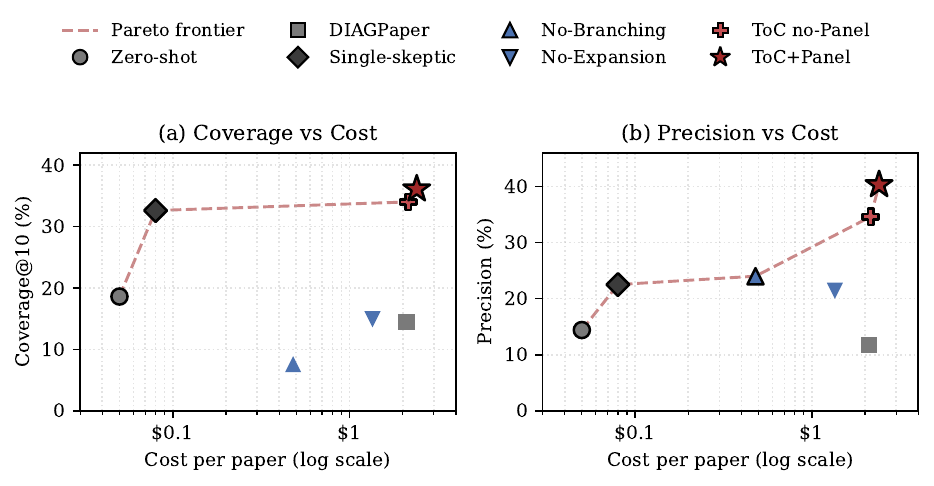}
\caption{Cost--quality trade-off.}
\label{fig:cost_pareto}
\end{figure*}

\noindent Figure~\ref{fig:cost_pareto} visualizes the cost--quality trade-off implied by Table~\ref{tab:compute} and Table~\ref{tab:main_results}. The dashed line in each panel marks the Pareto frontier: a method is Pareto-optimal if no other method achieves both lower cost and higher quality. On Coverage@10, \methodshort{}+Panel, ToC no-Panel, Single-skeptic, and Zero-shot are Pareto-optimal; on Precision, No-Branching joins the frontier. Most striking, \emph{Single-skeptic dominates DIAGPaper and No-Expansion on both axes}: at $\sim$\$0.08 per paper it already exceeds the Coverage and Precision of the two methods that cost \$1.35--\$2.10, indicating that those baselines do not convert their additional compute into useful supervision. The gap between Single-skeptic and \methodshort{}+Panel ($+3.5$ Coverage@10 and $+17.8$ Precision points at $\sim$30$\times$ the cost) quantifies the price of the framework's specialization plus debate machinery, and shows that the ToC headline gains come from \emph{architecture} rather than from raw API spend---if compute alone were the driver, the much cheaper Single-skeptic would already match ToC+Panel.
 
\section{Dataset Sample}
\label{app:dataset_sample}

Extending the benchmark description in \S\ref{sec:bench}, Table~\ref{tab:dataset_sample} shows a complete gold record from \bench{} for the DDSP paper. Each record pairs the paper's text with structured limitation annotations including the natural-language claim, its category and severity, the source type (OpenReview reviewer or citing paper), and the paper section where the supporting evidence appears. This example illustrates the typical profile: 4 of 5 limitations come from OpenReview (contemporaneous reviewer weaknesses), while 1 comes from a follow-up citation critique requiring knowledge of subsequent work.

\begin{table*}[t]
\small
\begin{tabular}{p{0.07\linewidth}p{0.05\linewidth}p{0.54\linewidth}p{0.12\linewidth}p{0.12\linewidth}}
\toprule
\textbf{Cat.} & \textbf{Sev.} & \textbf{Description} & \textbf{Source} & \textbf{Section} \\
\midrule
scope & major & DDSP autoencoder relies on accurate f0 estimation; failure modes when pitch tracking is unreliable are unacknowledged & OpenReview & Abstract \\
scope & minor & Title claims generality but experiments limited to monophonic pitched audio & OpenReview & Intro \\
repro. & minor & Frame rate of neural network relative to audio sample rate not reported & OpenReview & Experiments \\
method. & minor & Claims models are ``relatively small'' without complexity comparisons & OpenReview & Results \\
scope & minor & Differentiability constraint prevents reproducing commercial-grade FX dynamics & Citation & Abstract \\
\bottomrule
\end{tabular}
\caption{Complete \bench{} record for DDSP \citep{engel2020ddsp} (5 gold limitations).}
\label{tab:dataset_sample}
\end{table*}

\section{Case Studies}
\label{app:examples}
Building on the analysis in \S\ref{sec:analysis}, we present extended case studies demonstrating the full range of debate outcomes (valid, deflected), panel actions (endorse, merge, downgrade), and expansion behavior. Examples are drawn from real system outputs on multiple papers from the 100-paper held-out set.

\subsection{Full DDSP Output (All Branches)}
Table~\ref{tab:ddsp_full} shows the complete \method{} output for the DDSP paper across all five branches. This single paper illustrates every panel action type: endorse (4 claims), merge (2 claims with cross-category labeling), downgrade (2 claims with severity reduction), and deflection (1 branch terminated). The output comprises 8 valid limitations from 9 debate nodes---a precision-focused set where every surviving claim has been stress-tested through adversarial debate and cross-category panel review.

\begin{table*}[t]
\small
\begin{tabular}{cllllp{0.55\linewidth}}
\toprule
\textbf{D} & \textbf{Cat.} & \textbf{Verdict} & \textbf{Sev.} & \textbf{Panel} & \textbf{Claim} \\
\midrule
0 & Scope & valid & major & endorse & Abstract claims ``broad applicability'' but only monophonic pitched instruments tested \\
0 & Method. & valid & minor & endorse & WaveNet comparison uses parameter counts from different experimental setting \\
0 & Repro. & valid & minor & merge & No multi-seed results; FAD 2.79 vs 3.16 lacks statistical grounding \\
0 & Fairness & valid & minor & merge & Single-performer evaluation with no cross-performer diversity \\
0 & Theoretical & deflected & --- & --- & No formal proofs; engineering contribution \\
\midrule
1 & Scope & valid & minor & endorse & Harmonic-plus-noise cannot handle inharmonic/transient sources \\
1 & Scope & valid & major & endorse & ``High-fidelity generation'' misleading given monophonic-only validation \\
1 & Fairness & valid & minor & downgrade & Timbre transfer has no voice-characteristic analysis \\
1 & Fairness & valid & minor & downgrade & NSynth reports no per-instrument-family breakdowns \\
\bottomrule
\end{tabular}
\caption{Complete \method{} output for DDSP (9 nodes, 5 branches). Every panel action type is represented.}
\label{tab:ddsp_full}
\end{table*}

\subsection{Depth Expansion: Surface to Specific}
The Scope root (``broad applicability claimed, only monophonic tested'') expands into two children. \textbf{Child 1}: ``Harmonic-plus-noise assumes quasi-harmonic audio; drums and polyphony are architecturally excluded.'' The Advocate distinguishes the DDSP \emph{library} (general) from the \emph{model} evaluated (narrow); the skeptic sharpens the framing mismatch. Panel endorses scope/minor. \textbf{Child 2}: ``High-fidelity generation claimed generally but validated only on monophonic sources.'' Panel endorses scope/major---the strongest output claim. This shows how expansion converts one surface observation into two distinct, actionable sub-claims.

\subsection{Panel Merge: Cross-Category Labeling}
Reproducibility Skeptic: ``No multi-seed results; FAD 2.79 vs 3.16 lacks grounding.'' Panel assigns \textsc{merge} to \texttt{methodology} with cross-categories = [methodology, reproducibility]: ``Both a reproducibility concern (cannot assess variance) and methodology concern (close comparisons lack statistical support).''

\subsection{Deflection: Correct Abstention}
On ``Proving the Lottery Ticket Hypothesis,'' the Methodology Skeptic concedes: ``Purely theoretical paper with proofs, not experiments. Standard methodology concerns do not apply.'' On DDSP, the Theoretical Skeptic concedes: ``Engineering contribution without formal derivations.'' Both demonstrate correct per-category abstention maintaining precision.

\subsection{Non-DDSP: Scope Expansion on AUM Paper}
On ``Identifying Mislabeled Data using AUM,'' the Scope Skeptic raises: ``Claims `trivially compatible with any network' but only tests image classification.'' Expansion child: ``Does the core assumption (mislabeled $=$ low AUM) hold for NLP or tabular data where noise patterns differ?'' Ruled valid/minor. This shows the framework probing generalization assumptions beyond the tested domain.

\section{Evaluator Profiles}
\label{app:evaluators}
Following the evaluation protocol introduced in \S\ref{sec:eval_protocol}, the hybrid qualitative evaluation employs a four-member panel comprising two human evaluators and two LLM evaluators. The composition trades off depth (human evaluators bring methodological judgment and contextual understanding) and scale/consistency (LLM evaluators provide deterministic ratings on the full sample at low marginal cost):

\begin{itemize}[leftmargin=*, noitemsep, topsep=2pt]
\item \textbf{Evaluator 1 (human):} A fourth-year PhD student in computer science with eight publications; research expertise in NLP, multi-agent systems, and scientific document understanding. Familiar with the evaluation criteria from prior annotation experience on related tasks.

\item \textbf{Evaluator 2 (human):} A final-year undergraduate student in electrical engineering with two publications; research experience in machine learning applications. Trained on the evaluation guidelines with 10 practice examples before the main evaluation.

\item \textbf{Evaluator 3 (LLM):} GPT-4o, accessed via the OpenAI Chat Completions API at temperature 0.0 for deterministic ratings. Receives the same 1--5 Likert rubric as the human evaluators, formatted as the LLM-evaluator prompt template in Appendix~\ref{app:llm_likert}.

\item \textbf{Evaluator 4 (LLM):} Claude Opus 4.7, accessed via Anthropic API at temperature 0.0. Receives the identical prompt to Evaluator~3.
\end{itemize}

Both human evaluators are from the same organization. They participated voluntarily as part of the project and received no separate compensation. Both human evaluators were briefed on the task definition (unstated limitation extraction) and provided with the scoring rubric below. Neither human evaluator was involved in system development. Outputs were shuffled and anonymized so no evaluator could determine which method produced which output. The two LLM evaluators received the identical rubric programmatically (Appendix~\ref{app:llm_likert}) and were likewise blind to method identity, since prompts contain only the limitation text and the paper context, never the producing method.

\section{Output Comparison Across Methods}
\label{app:output_comparison}
To illustrate qualitative differences between the methods compared in \S\ref{sec:methods_compared}, we present the top-3 limitations produced by each of three methods on the same paper (DDSP) in Table \ref{tab:output_comparison}. 

\begin{table*}[t]
\small
\begin{tabular}{p{0.15\linewidth}p{0.80\linewidth}}
\toprule
\textbf{Method} & \textbf{Top-3 Limitations Produced for DDSP} \\
\midrule
\textbf{Zero-shot LLM} &
1. [scope] ``The paper only evaluates on monophonic audio'' \\
& 2. [scope] ``Relies on CREPE for pitch estimation'' \\
& 3. [methodology] ``Only solo violin evaluation with 13 minutes of data'' \\
\midrule
\textbf{Single-skeptic CoT} &
1. [scope] ``Experiments limited to monophonic, pitched instrument sounds; no polyphonic or non-pitched evaluation'' \\
& 2. [methodology] ``WaveRNN comparison uses numbers from different experimental conditions'' \\
& 3. [reproducibility] ``Training details for the solo violin model are underspecified (learning rate, batch size not reported)'' \\
\midrule
\textbf{ToC+Panel (ours)} &
1. [scope/major] ``Abstract claims `high-fidelity generation' and `broad applicability' but all experiments restricted to monophonic, quasi-harmonic sources---the Harmonic-plus-Noise architecture architecturally excludes drums, environmental sounds, and polyphony'' \\
& 2. [methodology/minor] ``Table 3 comparison against WaveNet autoencoder uses parameter counts from the original publication rather than a model retrained on the same solo-instrument data, making efficiency claims uncontrolled'' \\
& 3. [methodology+repro./minor] ``No multi-seed results or confidence intervals; close comparisons (FAD 2.79 vs 3.16) lack statistical grounding to support claimed improvements'' \\
\bottomrule
\end{tabular}
\caption{Side-by-side output comparison on DDSP. Zero-shot produces surface-level observations; Single-skeptic adds structure but remains broad; ToC+Panel produces specific, evidence-grounded claims with architectural reasoning and controlled-comparison analysis. Note how ToC+Panel's claim \#1 combines the scope concern (monophonic) with the \emph{architectural reason} (Harmonic-plus-Noise model), and claim \#3 carries a cross-category label from the panel.}
\label{tab:output_comparison}
\end{table*}

\section{Benchmark Dataset Specification}
\label{app:dataset_spec}

Extending the benchmark introduction in \S\ref{sec:bench}, \bench{} is distributed as a JSON file. Each record contains:

\begin{itemize}[leftmargin=*, noitemsep]
\item[(a)] \texttt{paper\_id}: canonical arXiv identifier
\item[(b)] \texttt{title}: paper title
\item[(c)] \texttt{abstract}: paper abstract
\item[(d)] \texttt{sections}: list of \{title, text\} section objects
\item[(e)] \texttt{limitations}: list of gold limitation records, each with:
  \begin{itemize}[noitemsep]
  \item \texttt{description}: 1--2 sentence limitation claim
  \item \texttt{category}: scope $|$ methodology $|$ theoretical $|$ reproducibility $|$ fairness
  \item \texttt{severity}: major $|$ minor
  \item \texttt{evidence\_in\_paper}: \{text, section\} quote from paper
  \item \texttt{source}: \{kind, citation, quote, url\}
  \end{itemize}
\item[(f)] \texttt{tier}: ``gold'' (OpenReview available) or ``silver'' (citation-only)
\end{itemize}

\end{document}